\documentclass{article}

\usepackage{microtype}
\usepackage{graphicx}
\usepackage{subfigure}
\usepackage{booktabs} 
\usepackage{bm}
\usepackage{amsmath}
\usepackage{amsthm}
\usepackage{amssymb}

\usepackage{hyperref}

\newtheorem{thm}{Theorem}[section] 

\usepackage[accepted]{icml2019}

\icmltitlerunning{Global Adaptive Generative Adjustment}

\begin{document}

\twocolumn[
\icmltitle{Global Adaptive Generative Adjustment}



\icmlsetsymbol{equal}{*}

\begin{icmlauthorlist}
\icmlauthor{Bin Wang}{equal,to,goo}
\icmlauthor{Xiaofei Wang}{equal,to}
\icmlauthor{Jianhua Guo}{to}
\end{icmlauthorlist}

\icmlaffiliation{to}{School of Mathematics and Statistics, Northeast Normal University, Changchun, China}
\icmlaffiliation{goo}{Changchun Institute of Optics, Fine Mechanics and Physics, Chinese Academy of Sciences, Changchun, Jilin Province, China}

\icmlcorrespondingauthor{Xiaofei Wang}{wangxf341@nenu.edu.cn}
\icmlcorrespondingauthor{Jianhua Guo}{jhguo@nenu.edu.cn}


\vskip 0.3in
]



\printAffiliationsAndNotice{\icmlEqualContribution} 

\begin{abstract}
Many traditional signal recovery approaches can behave well basing on the penalized likelihood.
However, they have to meet with the difficulty in the selection of hyperparameters or tuning parameters in the penalties.
In this article, we propose a global adaptive generative adjustment (GAGA) algorithm for signal recovery, in which 
multiple hyperpameters are automatically learned and alternatively updated with the signal.
We further prove that the output of our algorithm directly guarantees the consistency of model selection and signal estimate.
Moreover, we also propose a variant GAGA algorithm for improving the computational efficiency in the high-dimensional data analysis.
Finally, in the simulated experiment, we consider the consistency of the outputs of our algorithms, and compare our algorithms to other penalized likelihood methods: the Adaptive LASSO, the SCAD and the MCP.
The simulation results support the efficiency of our algorithms for signal recovery, and demonstrate that our algorithms outperform the other algorithms.

\end{abstract}

\section{Introduction}
\label{submission}
In the past two decades, signal recovery methods developed rapidly in the machine learning and statistics community. Much of the recent work push the boundaries of
our theoretical knowledge on the high-dimensional data analysis,
and offers a wide range of applications in the computer, biology and medicine fields. 
 Specially, several important approaches (\citealp{tibshirani:1996,chen:1998,fan:2001,candes:2005,zhao:2006}) have been developed for the rapid development in signal recovery;
see \citealp{hastie:2009} for an overview.

Though demonstrated effective in theoretical analysis,
the performance of the signal recovery relies on
an appropriate choice of the tuning parameters in the penalized likelihood.
The single tuning parameter selection has been studied in a series of works
using the BIC-type scoring criterions (\citet{wanghs2:2007,wanghs:2009,fanyy:2013,hui:2015}).
To conquer the oracle limitation of the LASSO (\citet{tibshirani:1996}),
\citet{zou:2006} proposed an adaptive version by introducing multiple hyperparamters for customizing a personalized shrinkage for each component in signal.
In the practical computation,
the adaptive LASSO considers the selection for a pair of hyperparameters
by the cross-validation (CV).
Other penalized likelihood methods: the SCAD (\citet{fan:2001}) and the MCP (\citet{zhang:2010})
also need to choose two hyperparameters over the two-dimensional grids using some scoring criteria.

To our best knowledge, there is no existing work accommodating the selection
directly for the multiple tuning parameters.
The traditional scoring search is not efficient any more since it incurs huge time-cost for the passive traveral in multi-dimensional threshold space.
Moreover, most theorectical properties on those penalized methods are based on the optimal solution of the objective function rather than the output of their algorithms. The gap between the optimal solution and the output could make the performance of their algorithms deviating from those expected theoretical properties.

Our present work contributes three novel points in the signal recovery:
Firstly, our proposed algorithm can alternatively update the signal and multiple hyperparameters in an active way. It provides an automatic learning of hyperparameters for signal recovery.
Secondly, we prove that the ouput of the algorithm enjoys both the consistency for model selection and signal estimate.
Thirdly, our proposed algorithm works in a concise form and performs well on both the error and the accuracy of the signal estimate in the computational aspect.

In our work, multiple tuning parameters are introduced for personalising the penalty on each component in the signal. Tuning parameters and the signal are alternatively updated
by a data-driven method, which we call as global adaptive generative adjustment (GAGA).
The GAGA updates multiple hyperparameters in a purposeful way. So it avoids the time-consuming scoring searching in the traditional hyperparameter selection.
By studying in detailed the iteration process of algorithm,
we prove that the output of the GAGA algorithm
directly possesses the consistency of both the model selection and the signal estimate.
Thus the output of the algorithm usually has a performance with a low error and a high accuracy when the sample size is large enough.
Furthermore, we propose another QR-decomposition version of the GAGA algorithm. This QR-version can improve the computational efficiency of the orignal one for the high-dimensional data analysis.
We illustrate the performance of our algorithms by several simulated experiments. Our algorithms outperform other penalized likelihood methods on the error and the accuracy of the signal estimate.
The time costs are also much lower of our algorithms than the others with the 10-fold Cross-Validation selection.

The rest of our paper is organized as follows.
In Section 2, we describe two versions of the Global Adaptive Generative Adjustment algorithm.
and present the theoretical guarantee of our algorithm.
In Section 3, we show the simulation results of our algorithms and
other popular penalized likelihood algorithms.
Finally, we give the conclusion on our algorithms in Section 4.
All the proof details are put into the Supplementary Material.

\section{Global Adaptive Generative Adjustment}

In this section, we describe the Global Adaptive Generative Adjustment (GAGA) algorithm,
and present the theoretical guarantees of the algorithm on a linear model with an orthogonal design matrix.
Furthermore, we propose a QR-decomposition version of the GAGA algorithm for improving its computational efficiency in the high-dimensional data analysis.

\subsection{A Start From A Simple Linear Model}

We start from a linear model
$\mathbf{y}=\mathbf{X}\bm{\beta}^*+\bm{\varepsilon}$,
where the true signal $\bm{\beta}^*=(\beta_1^*,\cdots,\beta_p^*)^T$, the noise $\bm{\varepsilon}\sim N(\mathbf{0},\sigma^2\mathbf{I})$ and
$\mathbf{I}$ is an identity matrix.
The recovery of the true signal $\bm{\beta}^*$ can be considered under a shrinkage framework with multiple tuning parameters. For getting a concise update form, we first assume that the variance $\sigma^2=1$. So $\bm{\varepsilon}\sim N(\mathbf{0},\mathbf{I})$. Specifically,
we take into account the ridge regression form
$$\frac{1}{2}\|\mathbf{y}-\mathbf{X}\bm{\beta}\|^2+\frac{1}{2}\sum_{j=1}^{p} \lambda_j{\beta_j}^2$$
with tuning parameters
$\lambda_j$, $j=1,\cdots,p$. The tuning parameter $\lambda_j$ customizes the amount of the penalty on the coefficient $\beta_j$. It can provide a personalized shrinkage on the coefficient. We introduce a global adaptive generative adjustment (GAGA) algorithm \ref{ASR} to recover a true signal $\bm{\beta}^*$.
In case that the variance is unknown,  it can be estimated by using the residual of the estimated signal. We will give the whole algorithm version with the estimated noise in the last part of this subsection.

In the algorithm, tuning parameters and the signal are alternatively updated
by a data-driven method.
So it avoids the time-consuming scoring searching in the traditional hyperparameter selection.
The inputs of this algorithm are the response vector $\mathbf{y}$, the design matrix $\mathbf{X}$, the iteration number $K$ and a constant $\alpha$.
The constant $\alpha$ can control the sparsity of the signal estimate. We set $\alpha=2$ in the whole simulation experiment part.
The output of this algorithm is the signal estimate
$\hat{\bm{\beta}}=\text{GAGA}(\mathbf{y},\mathbf{X},K,\alpha)$.

As shown in Algorithm 1, the estimate on the signal $\bm{\beta}^*$ is updated by a ridge regression form in Line 3. This regression relies on a diagonal matirx $\bm{\Lambda}$.
Its diagonal elements are those personalized tuning parameters, which are updated in Line 4.
The tuning parameter vector $\bm{\lambda}$
obtains a global adaptive update form in Line 4
depending on the data $\mathbf{X}$ and $\mathbf{y}$.
Furthermore,
the vector $\bm{\lambda}$ can
provide a generative adjustment
for the signal estimate in the next iteration shown in Line 3.
After $K$ iterations, we judge a hard truncation condition in Line 9. The condition determines that the estimated coefficient $\hat{\beta}_j^{(K)}$
is shrinked to zero or not.

\begin{algorithm}[h]
	\caption{Global Adaptive Generative Adjustment (GAGA)}
	\label{ASR}
	\textbf{Input:} Response vector $\mathbf{y}$,  design matrix $\bm{X}$,  iteration number $K$, growth factor $\alpha$. \\
	\textbf{Output:} The signal estimate $\hat{\bm{\beta}}^*=(\hat{\beta}_1^*,\cdots,\hat{\beta}_p^*)$. \\
	\textbf{Main Procedure:}
	\begin{algorithmic}[1]
		\STATE Hyperparameter vector $\bm{\lambda}^{(1)}=(\lambda_1^{(1)},\cdots,\lambda_p^{(1)})^T=\mathbf{0}$.
		\FOR {$k=1,2,\ldots K-1$}
		\STATE $\hat{\bm{\beta}}^{(k)}=(\mathbf{X}^T\mathbf{X}+\bm{\Lambda}^{(k)})^{-1}\mathbf{X}^T\mathbf{y}$ where $\bm{\Lambda}^{(k)}=diag(\bm{\lambda}^{(k)})$.
		\STATE $\lambda_j^{(k+1)}=\frac{\alpha}{(\hat{\beta}_j^{(k)})^2+((\mathbf{X}^T\mathbf{X}+\bm{\Lambda}^{(k)})^{-1})_{jj}}$, $j=1,\cdots,p$.
		\ENDFOR
		\STATE
		$\bm{\lambda}^{(K)}=\bm{\lambda}^{(K)}/\alpha$.
		\STATE
		$\hat{\bm{\beta}}^{(K)}=(\mathbf{X}^T\mathbf{X}+\bm{\Lambda}^{(K)})^{-1}\mathbf{X}^T\mathbf{y}$ where $\bm{\Lambda}^{(K)}=diag(\bm{\lambda}^{(K)})$.
		\FOR {$j=1,2,\ldots p$}
		\IF {$(\hat{\beta}_j^{(K)})^2 \leq ((\mathbf{X}^T\mathbf{X})^{-1})_{jj}-((\mathbf{X}^T\mathbf{X}+\bm{\Lambda}^{(K)})^{-1})_{jj}$}
		\STATE $\hat{\beta}_j^* \leftarrow 0$.
		\ELSE
		\STATE $\hat{\beta}_j^* \leftarrow \hat{\beta}_j^{(K)}$.
		\ENDIF
		\ENDFOR

	\end{algorithmic}
\end{algorithm}

In the following part of this subsection,
we show the theorectical guarantees on the output of the GAGA algorithm when the design matrix is column orthogonal. Those guarantees illustrate that 
the GAGA algorithm can provide an efficient
estimate on the signal under some conditions.
The empirical performance of the GAGA will
be shown in Section 3.

We denote $\mathcal{Q}$ as the subscript set $\{j|\beta_j^*\neq 0\}$ for non-zero components of signal. 
Assume that the design matrix $\mathbf{X}=(\textbf{x}_1,\cdots,\textbf{x}_p)$ is column orthogonal.
That is $\mathbf{X}^T\mathbf{X}=diag((a_1,\cdots,a_p))$.
And further assume that the condition number  $\kappa=\frac{\max_ja_j}{\min_ja_j}$ is bounded.
Since the column orthogonality of $\mathbf{X}$, the diagonal element
$((\mathbf{X}^T\mathbf{X}+\bm{\Lambda}^{(k)})^{-1})_{jj}=(a_j+\lambda_j^{(k)})^{-1}$.
So the update of $\lambda_j^{(k+1)}$ in Line 4 of the GAGA algorithm can be computed by $ \frac{\alpha(\lambda_j^{(k)}+a_j)^2}{\lambda_j^{(k)}+a_j+z_j}$ where $z_j=(\textbf{x}_j^T\textbf{y})^2$. Moreover, the hard thresholding condition in Line 9 becomes
$(\hat{\beta}_j^{(K)})^2\leq a_j^{-1}-(a_j+\lambda_j^{(K)})^{-1}$.
Unlike the conventional discussion on the optimum of the penalized likelihood,
we prove that the output of our GAGA algorithm
directly satisfies  the consistency of model selection and signal estimate.

For any convergent subsequence $\{\bm{\lambda}^{(k_l)}\}_l$ of the updated hyperparameter sequence $\{\bm{\lambda}^{(k)}\}_k$,
denote $\bm{\lambda}^\infty$ as the limit $\lim\limits_{l\rightarrow\infty}\bm{\lambda}^{(k_l)}$ (Components of the limit are allowed to be the infinity). 
Let $\bm{\lambda}^*=\bm{\lambda}^{\infty}/\alpha$, $\bm{\Lambda}^*=diag(\bm{\lambda}^*)$, and $\hat{\bm{\beta}}=(\mathbf{X}^T\mathbf{X}+\bm{\Lambda}^*)^{-1}\mathbf{X}^T\mathbf{y}$. Furthermore, let $\hat{\bm{\beta}}^*$ be the personalized thresholding of $\hat{\bm{\beta}}$ in the GAGA algorithm.
Theorem \ref{beta=0} illustrates the effectiveness of the hard truncation in Line 10 of the GAGA algorithm.
If the true coefficient is zero, the hard truncation happens with a high probability when the sample size is large enough. It means that the zero-coefficient position can be correctly detected with a high probability.

\begin{thm}\label{beta=0}
	Assume that the design matrix is column orthogonal.
	We have that the probability for personalized thresholding in $\mathcal{Q}^c$
	\begin{equation*}
		\begin{aligned}
			&\mathbb{P}(\bigcap\limits_{j\in\mathcal{Q}^c}\{(\hat{\beta}_j)^2\leq a_j^{-1}-(a_j+\lambda_j^*)^{-1}\})\\
			\geq
			&1-\exp(-\frac{1}{2}(\sqrt{\alpha}+\sqrt{\alpha-1})^2+\log(p-q))\\
		\end{aligned}
	\end{equation*} 
	Moreover, for any $0<\eta<1$, the probability for no personalized thresholding in $\mathcal{Q}$
	\begin{equation*}
		\begin{aligned}
			&\mathbb{P}(\bigcap\limits_{j\in\mathcal{Q}}\{(\hat{\beta}_j)^2> a_j^{-1}-(a_j+\lambda_j^*)^{-1}\})\\
			\geq
			&1-	2\exp(-\frac{1}{2}(\min\limits_{j\in \mathcal{Q}}a_j^{1/2}|\beta_j|-(\sqrt{\alpha}+\sqrt{\alpha-1}))^2+\\
			&\log(q)) -\exp(-\frac{1}{2}((1-\eta)\min\limits_{j\in \mathcal{Q}}a_j^{1/2}|\beta_j^*|-\\
			&(\sqrt{\alpha}+\sqrt{\alpha-1}))^2+\log(q))
		\end{aligned}
	\end{equation*}
when the inequalities $\eta^2\min\limits_{j\in \mathcal{Q}}a_j{\beta_j^*}^2\geq \frac{1}{\sqrt{\alpha}+\sqrt{\alpha-1}-1}$ and $(1-\eta)\min\limits_{j\in \mathcal{Q}}a_j^{1/2}|\beta_j^*|-(\sqrt{\alpha}+\sqrt{\alpha-1})\geq 0$ hold.
\end{thm}

The following Theorem \ref{consistency} provides an error expectation bound for the output $\hat{\bm{\beta}}^*$ of the GAGA algorithm. Let the event
$E=\{z_j <(\sqrt{\alpha}+\sqrt{\alpha-1})^2a_j,j\in\mathcal{Q}^c\}
$,
the event
$F=\{\max\limits_{j\in\mathcal{Q}}\lambda_j^*{\beta_j^*}^2\leq\frac{1}{\eta^2}\}$
where $0<\eta<1$, and the condition number $\kappa=\frac{\max_ja_j}{\min_ja_j}$.

\begin{thm}\label{consistency}
Assume that the design matrix is column orthogonal, and the condition number $\kappa$ is bounded.
	We have that 
	\begin{equation*}
	\begin{aligned}
		&\mathbf{E}\|\hat{\bm{\beta}}^*-\bm{\beta}^*\|\\
		\leq & \sqrt{\frac{q}{\min\limits_{j\in \mathcal{Q}}a_j}}(1+\sqrt{\kappa}+\frac{1}{\eta^2\min\limits_{j\in \mathcal{Q}}a_j^{1/2}|\beta_j^*|})+\\
		& (E\|\hat{\bm{\beta}}\|^2)^{1/2}(2(\mathbb{P}(E^c))^{1/2}+(\mathbb{P}(F^c))^{1/2})
		+ \\
		& \|\bm{\beta}^*\|(\mathbb{P}(E^c)+\mathbb{P}(F^c)).\\
	\end{aligned}
\end{equation*}
	Moreover, we have that 
	$$	\mathbb{E}\|\hat{\bm{\beta}}^*-\bm{\beta}^*\|\xrightarrow{\alpha\rightarrow\infty,\min\limits_{j\in \mathcal{Q}}a_j\rightarrow\infty}0$$
	if further $q=o(\min\limits_{j\in \mathcal{Q}}a_j)$, $\log(p)+\log(\frac{p}{\min\limits_{j}a_j}+\|\bm{\beta}^*\|^2)=o(\alpha)$ and $\alpha=o(\min\limits_{j\in \mathcal{Q}}a_j{\beta_j^*}^2)$.
\end{thm}
All the techinial details can be found in the Supplementary Material.
In the following subsection,
we will propose a QR-decomposition version of the GAGA algorihm
for the high-dimensional data analysis.
Moreover, we will find that this new version is compatible with our developed theory under the orthogonal design assumption.

In case that the variance is unknown, we actually can estimate the variance in each iteration by the residual of the estimated signal:
\begin{equation}
(\sigma^2)^{(k+1)} = (\mathbf{X}\hat{\bm{\beta}}^{(k)}-\mathbf{y})^T(\mathbf{X}\hat{\bm{\beta}}^{(k)}-\mathbf{y})/N
\end{equation}
or
\begin{equation}\label{gaganoise}
\begin{aligned}
(\sigma^2)^{(k+1)} =&\frac{1}{N} \bigg(\mathbf{y}^T\mathbf{y}-2\hat{\bm{\beta}}^{(k) T}\mathbf{X}^T\mathbf{y}+\\
&\text{tr}\big( ( \hat{\bm{\beta}}^{(k) T} \hat{\bm{\beta}}^{(k)} +\mathbf{D}^{(k)})\mathbf{X}^T\mathbf{X} \big)\bigg)
\end{aligned}
\end{equation}
where $\mathbf{D}^{(k)}=(\sigma^2)^{(k)}(\mathbf{X}^T\mathbf{X}+(\sigma^2)^{(k)}\bm\Lambda^{(k)})^{-1}$.
Correspondingly, the hard truncation condition turns into the inequality:
\begin{equation}\label{truncwithnoise}
	\begin{aligned}
		(\hat{\beta}_j^{(K)})^2 <& (\sigma^2)^{(K)}((\mathbf{X}^T\mathbf{X})^{-1})_{jj}-\\
			&(\sigma^2)^{(K)}((\mathbf{X}^T\mathbf{X}+(\sigma^2)^{(K)}\bm{\Lambda}^{(K)})^{-1})_{jj}
	\end{aligned}
\end{equation}

\begin{algorithm}[!h]
	\caption{Global Adaptive Generative Adjustment (GAGA) with Estimated Variance}
	\label{GAGA}
	\textbf{Input:}  the response vector $\mathbf{y}$, the design matrix $\bm{X}$, the iteration number $K$, the constant $\alpha$. \\
	\textbf{Output:} the signal estimate $\hat{\bm{\beta}}^*$. \\
	\textbf{Main Procedure:}
	\begin{algorithmic}[1]
		\STATE $\bm{\lambda}^{(1)}=(\lambda_1^{(1)},\cdots,\lambda_p^{(1)})^T=\mathbf{0}$, $(\sigma^2)^{(1)}=1$.
		\FOR {$k=1,\ldots ,K-1$}
		\STATE $\hat{\bm{\beta}}^{(k)}=(\mathbf{X}^T\mathbf{X}+(\sigma^2)^{(k)}\bm{\Lambda}^{(k)})^{-1}\mathbf{X}^T\mathbf{y}$ where 	$\bm{\Lambda}^{(k)}=diag(\bm{\lambda}^{(k)})$. 
		\STATE $\lambda_j^{(k+1)}=\frac{\alpha}{(\hat{\beta}_j^{(k)})^2+(\sigma^2)^{(k)}((\mathbf{X}^T\mathbf{X}+(\sigma^2)^{(k)}\bm{\Lambda}^{(k)})^{-1})_{jj}}$ where $j=1,\cdots,p$.
		\STATE Update $(\sigma^2)^{(k+1)}$ by the equation (\ref{gaganoise}).
		\ENDFOR
	    \STATE  $\bm{\lambda}^{(K)}=\bm{\lambda}^{(K)}/\alpha$
	    \STATE $\hat{\bm{\beta}}^{(K)}=(\mathbf{X}^T\mathbf{X}+(\sigma^2)^{(K)}\bm{\Lambda}^{(K)})^{-1}\mathbf{X}^T\mathbf{y}$ where $\bm{\Lambda}^{(K)}=diag(\bm{\lambda}^{(K)})$.
			\FOR {$j=1,2,\ldots p$}
		\IF {the inequality (\ref{truncwithnoise}) holds,}
		\STATE $\hat{\beta}_j^* \leftarrow 0$.
		\ELSE
		\STATE $\hat{\beta}_j^* \leftarrow \hat{\beta}_j^{(K)}$.
		\ENDIF
		\ENDFOR
	\end{algorithmic}
\end{algorithm}

\subsection{Another Version of the GAGA Algorithm}
Since the GAGA computes the matrix inversion in each iteration, the efficency of the  algorithm may be limited for the high-dimensional data. So
we further propose a variant version GAGA\_QR for dealing with this problem.
This version first roughly estimates the signal vector by the least-square solution $\bm{\gamma}=(\mathbf{X}^T\mathbf{X})^{-1}\mathbf{X}^T\mathbf{y}$. And then sort the coefficients of $\bm{\gamma}$ in a decreasing absolute value ordering. When the sample size is large enough,
the sorted estimate $\bm{\gamma}$ could be an apropriate approximation of the true signal $\bm{\beta}^*$, whose zero coefficients are arranged
in the tail part. Permutate the columns of the design matrix $\mathbf{X}$ by $\mathbf{X}_{\text{new}}=\mathbf{X}\mathbf{P}$, where $\mathbf{P}$ is a permutation matrix  according to the rearrangement of those coefficients in $\bm{\gamma}$.
Do the QR-decomposition on the permuted design matrix $\mathbf{X}_{\text{new}}=\mathbf{Q}\mathbf{R}$, where $\mathbf{Q}$ is a column orthogonal matrix and $\mathbf{R}$ is an upper triangular matrix.
Under the QR-decomposition, the original linear model can be viewed as $\mathbf{y}=\mathbf{Q}\bm{\mathbf{\theta}}+\bm{\epsilon}$,
where $\bm{\theta}=\mathbf{R}\bm{\beta}^*$.
Furthermore, we use the GAGA algorihtm for the response vector $\mathbf{y}$ and the new design matrix $\mathbf{Q}$.
The final signal estimate is obtained by $\mathbf{P}*\mathbf{R}^{-1}*\text{GAGA}(\mathbf{y},\mathbf{Q},K,\alpha)$,
where $K$ is the iteration number and $\alpha$ is a constant.

\begin{algorithm}[h]
	\caption{Global Adaptive Generative Adjustment Using QR-Decomposition (GAGA\_QR)}
	\label{GAGAQR}
	\textbf{Input:} the response vector $\mathbf{y}$, the design matrix $\bm{X}$, the iteration number $K$, the constant $\alpha$ \\
	\textbf{Output:} the signal estimate $\hat{\bm{\beta}}^*$ \\
	\textbf{Main Procedure:}
	\begin{algorithmic}[1]
		\STATE Compute the least-square estimate\\ $\bm{\gamma}=(\mathbf{X}^T\mathbf{X})^{-1}\mathbf{X}^T\mathbf{y}$.
		
		\STATE Sort elements of $\bm{\gamma}$ in a decreasing absolute value order. 
		\STATE Rearrange the columns of $\mathbf{X}$ by $\mathbf{X}_{\text{new}}=\mathbf{X}\mathbf{P}$, where $\mathbf{P}$ is a permutation matrix according to the decreasing absolute value order of $\bm{\gamma}$.
		\STATE Use the QR-decomposition for $\mathbf{X}_{\text{new}}$. Let $\mathbf{X}_{\text{new}}=\mathbf{Q}\mathbf{R}$ where $\mathbf{Q}$ is a column orthogonal matrix and $\mathbf{R}$ is an upper triangular matrix.
		\STATE Compute $\bm{\theta}$=GAGA(\textbf{y},\textbf{Q},$K$,$\alpha$) by using the GAGA algorithm
		\STATE The estimate signal  $\hat{\bm{\beta}}^*=\mathbf{P}\mathbf{R}^{-1}\bm{\theta}$.

	\end{algorithmic}
\end{algorithm}

Since the matrix $\mathbf{R}$ is an upper triangular matrix, so does the matrix $\mathbf{R}^{-1}$.
If the estimate $\bm{\gamma}$ captures the correct rearrangement matrix $\mathbf{P}$ such that the true signal $\bm{\beta}^*$ with zero coefficients in its tail, the linear transform $\bm{\theta}=\mathbf{R}\bm{\beta}^*$ maintains the sparse tail part as $\bm{\beta}^*$ since $\mathbf{R}$ is an upper triangular matrix.
Moreover, the inverse linear tranform $\mathbf{R}^{-1}\bm{\theta}$ also keeps the sparse structure in the tail part as $\bm{\theta}$.
So the GAGA\_QR is efficient once the GAGA successfully finds those zero coefficients in the signal.

Note that in the GAGA\_QR algorithm, the inversion of matrix $\mathbf{X}^T\mathbf{X}$ is only computed once in Line 1.
Since the column orthogonality of $\mathbf{Q}$,
the inversion computation is easy in the GAGA algorithm
with the inputs $\mathbf{y}$, $\mathbf{Q}$, $K$ and $\alpha$.
Though another matrix inversion is asked in Line 6 of the GAGA\_QR,
the computation is also easy since the matrix $R$ is an upper traingular matrix.
The computation on
$(\mathbf{X}^T\mathbf{X})^{-1}$ is only doned once in the GAGA\_QR algorithm, while the GAGA algorithm has to compute the inversion matrix $(\mathbf{X}^T\mathbf{X}+\bm{\Lambda}^k)^{-1}$ in each iteration.
So the GAGA\_QR algorithm takes less time costs than the GAGA algorithm, especially when we cope with a high-dimensional design matrix $\mathbf{X}$ with a large number of columns. 
The experiment on the comparison between the GAGA and the GAGA\_QR will be shown in the next section.

\section{Simulation}
In this section, we do experiments to show
the performances of our algorithms on the simulated data. We first compare our algorithms to the SCAD (\cite{fan:2001}), the adaptive LASSO (\cite{zou:2006}) and the MCP (\cite{zhang:2010}) on two models,
whose sparse structures are from
\cite{tibshirani:1996}.
We further design another model for testing them on high-dimensional data.
And then, we demonstrate their performances on the asymptotic property of those estimates as the sample size increases.
Finally, we show the time costs of our algorithms on the high-dimensional data analysis.

Their performances
are evaluated by the error (ERR) and the accuracy (ACC).
The ERR is computed with $\|\hat{\bm{\beta}}-\bm{\beta}^*\|$
on the value difference between the estimated and true one.
The ACC is defined by the ratio $\frac{\text{True positives}+\text{True negatives}}{Positives+Negatives}$
of correctly finding the positions of zeros and non-zeros in the true signal $\bm{\beta}^*$.

The adaptive LASSO, the SCAD and the MCP are executed in the R with the ncvreg library \cite{ncvreg}. All the experiments in R adopt the OpenBlas for performing basic vector and matrix operations.
For these penalized likelihood algorithms, extra tuning parameters are needed to
estimate the parameter and regularize the model selection.
We consider to set $100$ values for the tuning parameter in the adaptive LASSO, the MCP and the SCAD respectively, and use the $10$-fold Cross-Validation to select the appropriate tuning parameters for them. We further demonstrate their performance by averaging the ERR and the ACC on each value point of the tuning parameter. In the simulation, the iteration number $K$ is set to $50$ and the constant $\alpha=2$ for the GAGA and the GAGA\_QR.

\begin{figure}[t]
	\centering
	\subfigure[Performances on ERR] {\includegraphics[height=2in,width=2.5in]{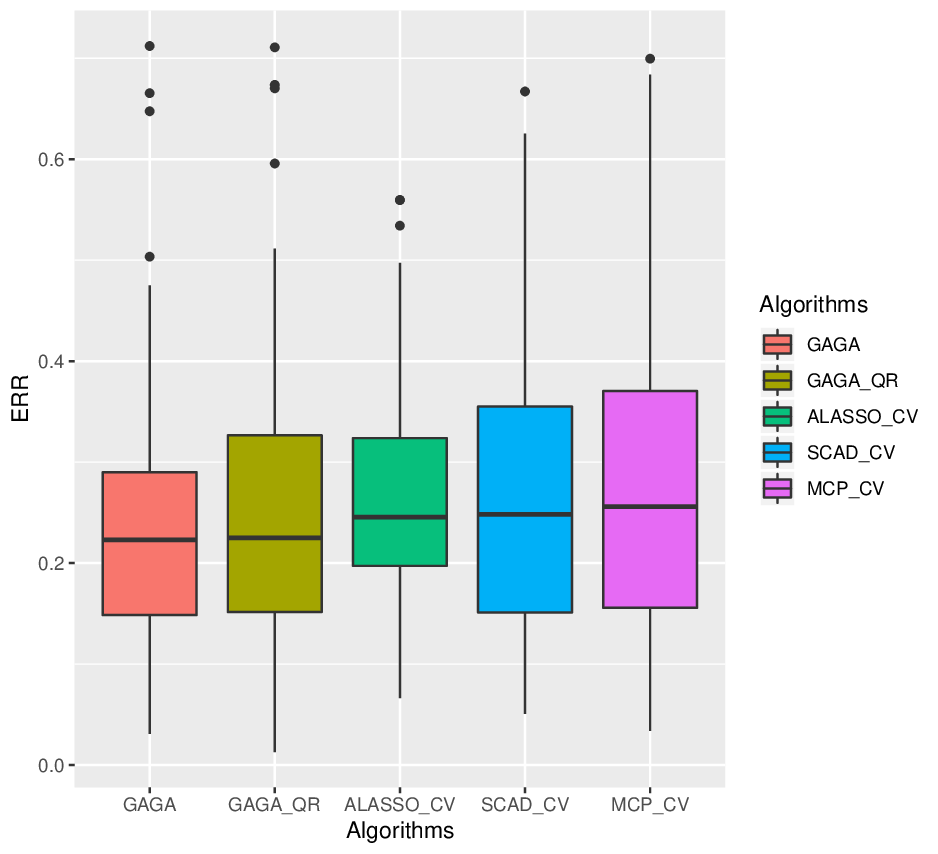}}
	\subfigure[Performances on ACC] {\includegraphics[height=2in,width=2.5in]{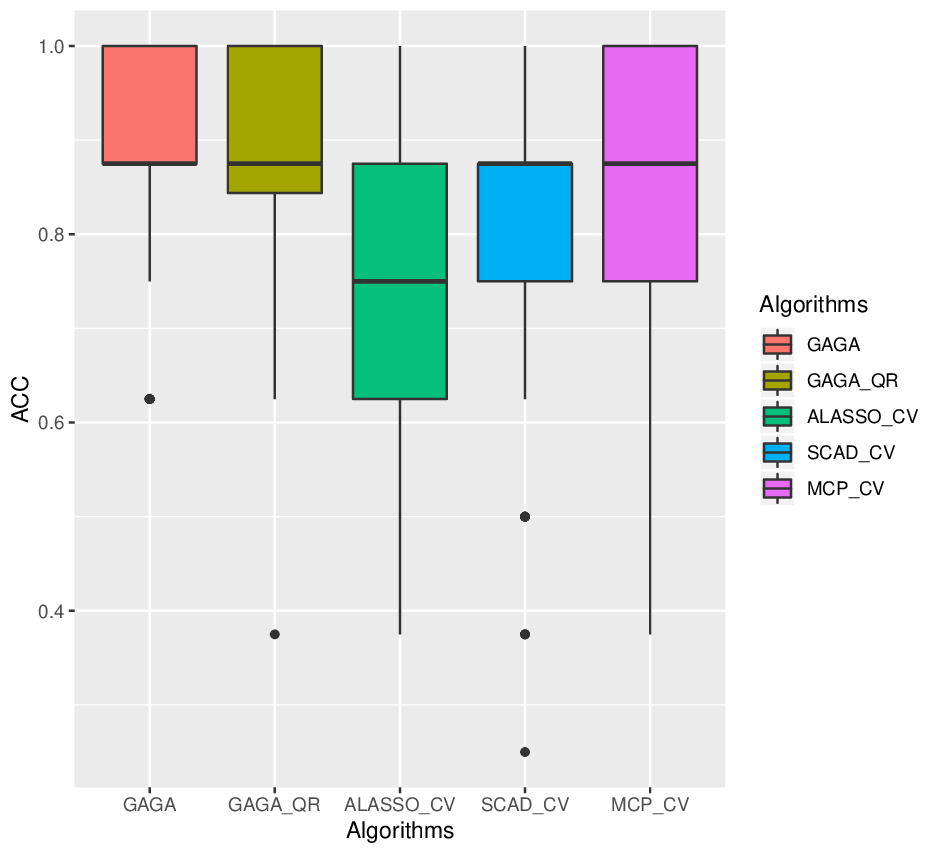}}
	\caption{Consider the ERR and the ACC seperately on Model $1$.}\label{experiment}
\end{figure}

\begin{figure}[htb]
	\centering
	{\includegraphics[height=3in,width=3.5in]{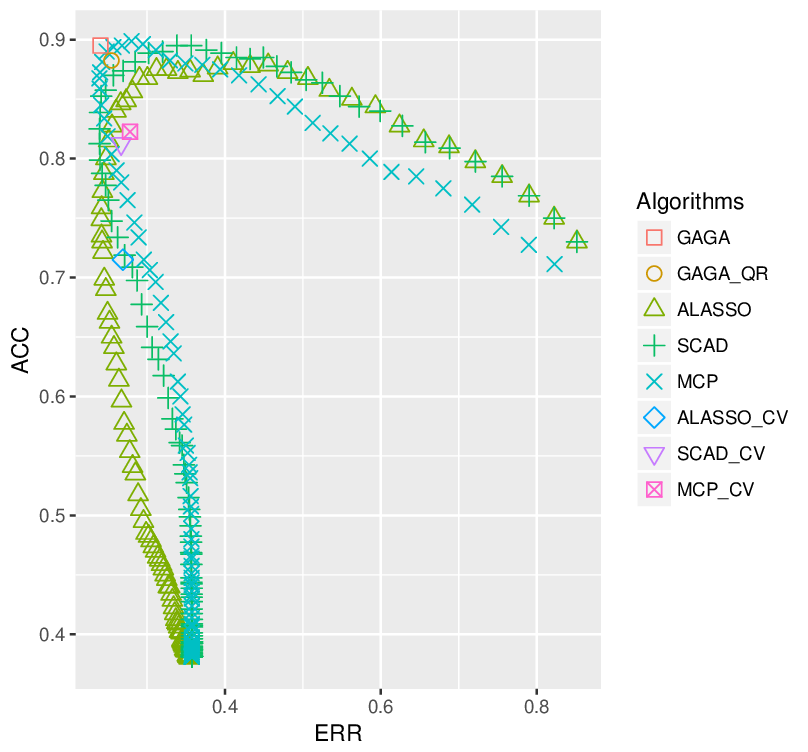}}
	\caption{Consider the ERR and the ACC comprehensively on Model $1$.}\label{experiment_ERR_ACC}
\end{figure}

For the Model $1$, $100$ data sets are simulated and each consists of $100$ observations
from the linear model $y=\mathbf{x}^T\bm{\beta}^*+\varepsilon$.
The noise $\varepsilon$ is a standard normal random variable.
The correlation between $x_i$ and $x_j$ is $\rho^{|i-j|}$ with $\rho=0.5$.
The true signal is set to $\bm{\beta}^*=(\beta_1^*,\beta_2^*,0,0,\beta_3^*,0,0,0)^T$.
The non-zero coefficients $\beta_1^*,\beta_2^*,\beta_3^*$ are randomly generated from $U(0,1)$.
The comparisons between ours and those algorithms are shown in Figure \ref{experiment}. For both the ERR and the ACC, the GAGA algorithm outperforms the adaptive LASSO, the SCAD and the MCP with the $10$-fold CV selection. Furthermore, we plot Figure \ref{experiment_ERR_ACC} to comprehensively demonstrate the performance of algorithms on the ERR and the ACC.
Each point in Figure \ref{experiment_ERR_ACC} represents an average ACC and ERR of $100$ data sets. The performances of the GAGA, the GAGA\_QR, the ALASSO\_CV, the SCAD\_CV and the MCP\_CV are characterized by five points. For the adaptive LASSO, the SCAD and the MCP, each algorithm have $100$ points representing the average performance on $100$ values of the tuning parameter. The two points for our algorithms are in the top left corners of Figure \ref{experiment_ERR_ACC}.
They perform better than other algorithms when comprehensively considering the ERR and the ACC.
Though the SCAD\_CV and the MCP\_CV behave well, they do not reach the best ones, which can be obtained by going over the values of the tuning paramter when knowing the ERR and the ACC. Since the true signal is unknown, it is not practical to select the tuning parameter by computing the ERR and the ACC, while our algorithms automatically learn all the tuning parameters and achieve a signal estimate with a low error and a high accuracy.

\begin{figure}[!htb]
	\centering
	\subfigure[Performances on ERR] {\includegraphics[height=2in,width=2.5in]{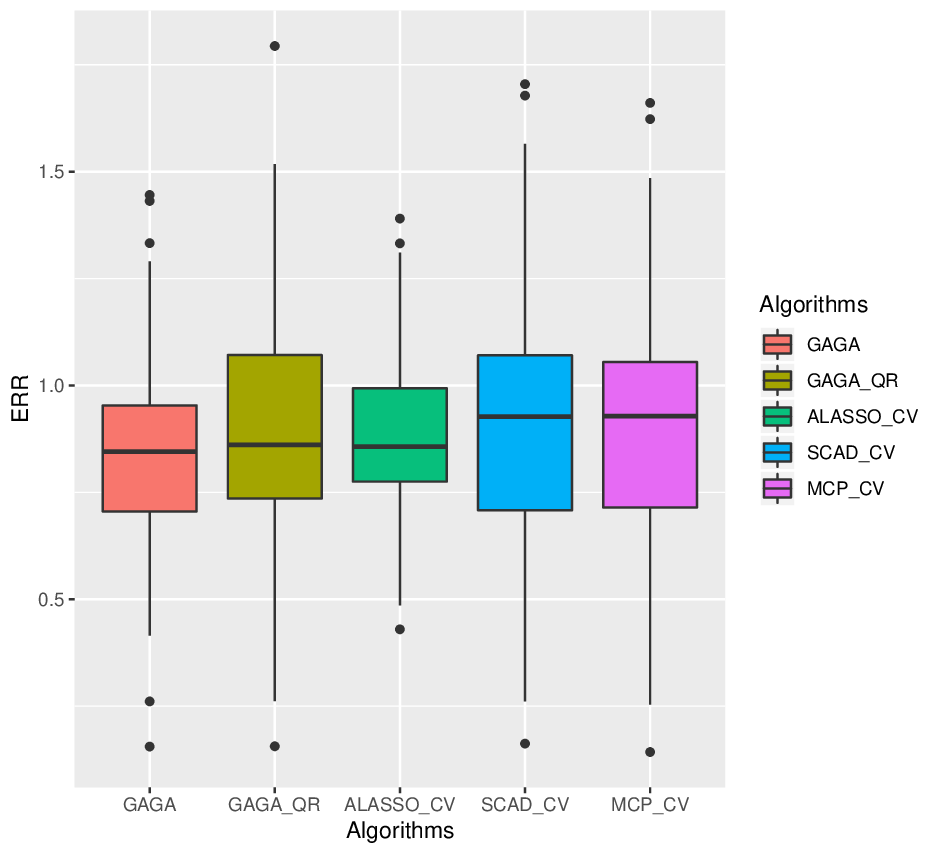}}
	\subfigure[Performances on ACC] {\includegraphics[height=2in,width=2.5in]{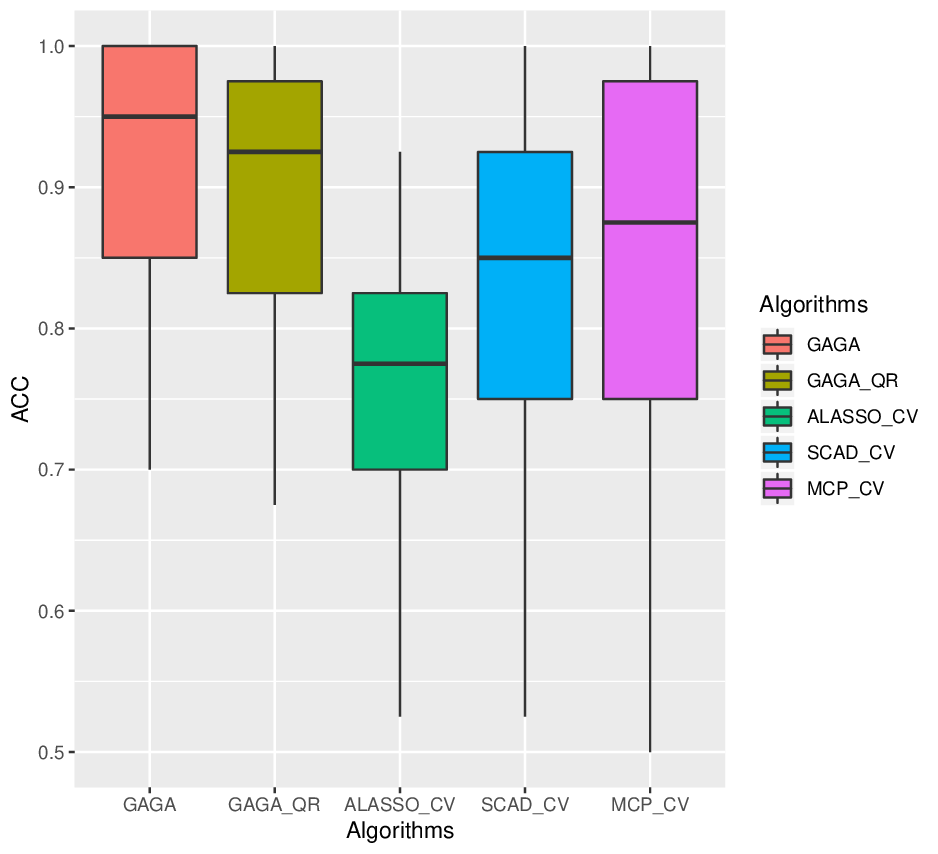}}
	\caption{Consider the ERR and the ACC seperately on Model $2$.}\label{experiment2}
\end{figure}

\begin{figure}[!htb]
	\centering
	{\includegraphics[height=3in,width=3.5in]{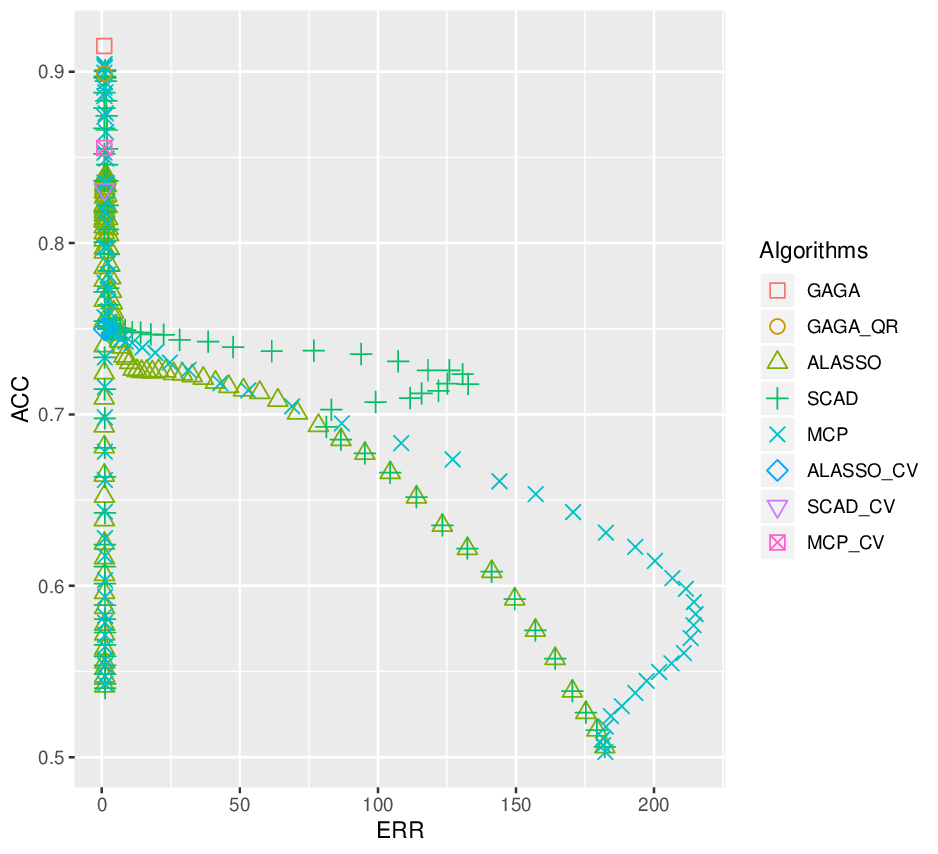}}
	\caption{Consider the ERR and the ACC comprehensively on Model  $2$.}\label{experiment2_ERR_ACC}
\end{figure}

For the Model $2$, $100$ data sets are simulated, and each had $100$ observations
from the linear model $y=\mathbf{x}^T\bm{\beta}^*+\varepsilon$.
The noise $\varepsilon$ is a standard normal random variable.
For any $i\neq j$, $x_i$ and $x_j$ have a pairwise correlation of $0.5$.
The true coefficient vector $$\bm{\beta}^*=(0,\cdots,0,\beta_1^*,\cdots,\beta_1^*,0,\cdots,0,\beta_2^*,\cdots,\beta_2^*)^T,$$
there being $10$ repeats in each block.
The non-zero components $\beta_1,\beta_2$ are randomly generated from $\beta_1\sim U(0,1)$ and $\beta_2\sim U(10,100)$ respectively.
As shown in Figure \ref{experiment2},
for the ACC, the GAGA and the GAGA\_QR outperform the adaptive LASSO, the SCAD and the MCP with the $10$-fold CV selection.
For the ERR, the GAGA performs better than the adaptive LASSO, the SCAD and the MCP.
As shown in Figure \ref{experiment2_ERR_ACC}, our algorithms also perform better than other penalized likelihood algorithms with the CV when comprehensively considering the ERR and the ACC.
We also set $100$ values of the tuning parameter in the adaptive LASSO, the MCP and the SCAD respectively.
So there are $100$ points in Figure \ref{experiment2_ERR_ACC} for each penalized likelihood algorithm, and each point represents an average ACC and ERR of algorithms on $100$ data sets.
The performance of the GAGA is still beyond those penalized likelihood algorithms with the hyperparameter selection in $100$ values
even if knowing the true ERR and ACC.

\begin{figure}[h]
	\centering
	\subfigure[Performances on ERR] {\includegraphics[height=1.7in,width=2.5in]{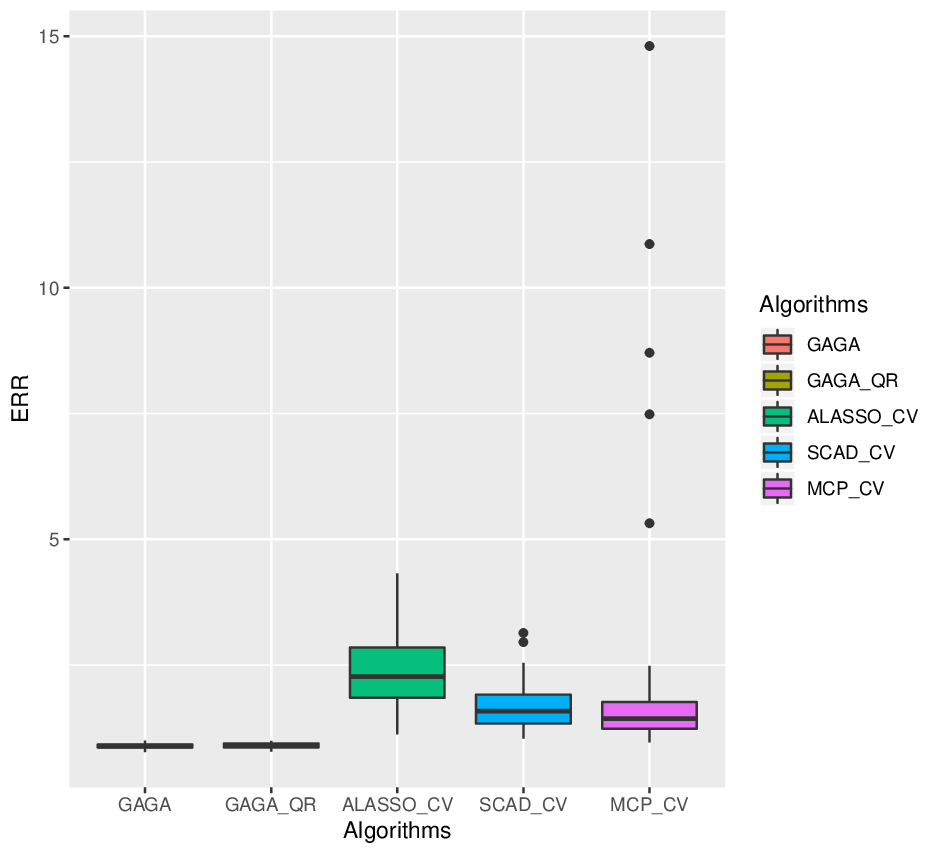}}
	\subfigure[Performances on ACC] {\includegraphics[height=1.8in,width=2.5in]{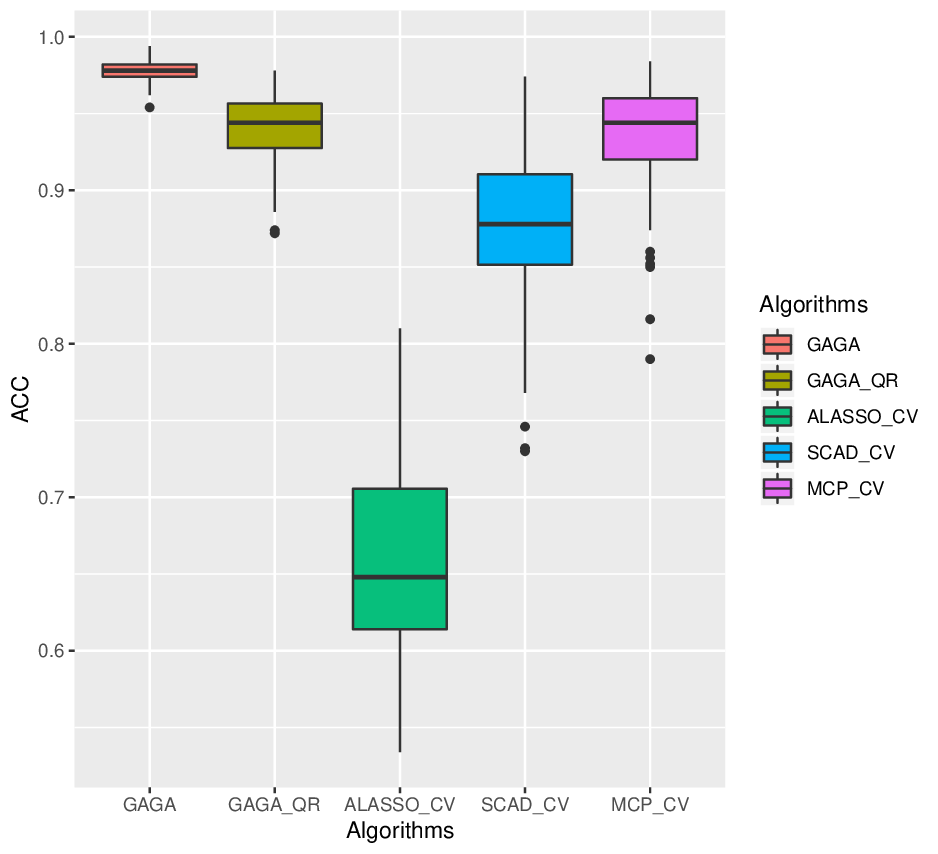}}
	\caption{Consider the ERR and the ACC seperately on the high-dimensional data.}\label{experiment3}
\end{figure}

\begin{figure}[!htbp]
	\centering
	{\includegraphics[height=3in,width=3.5in]{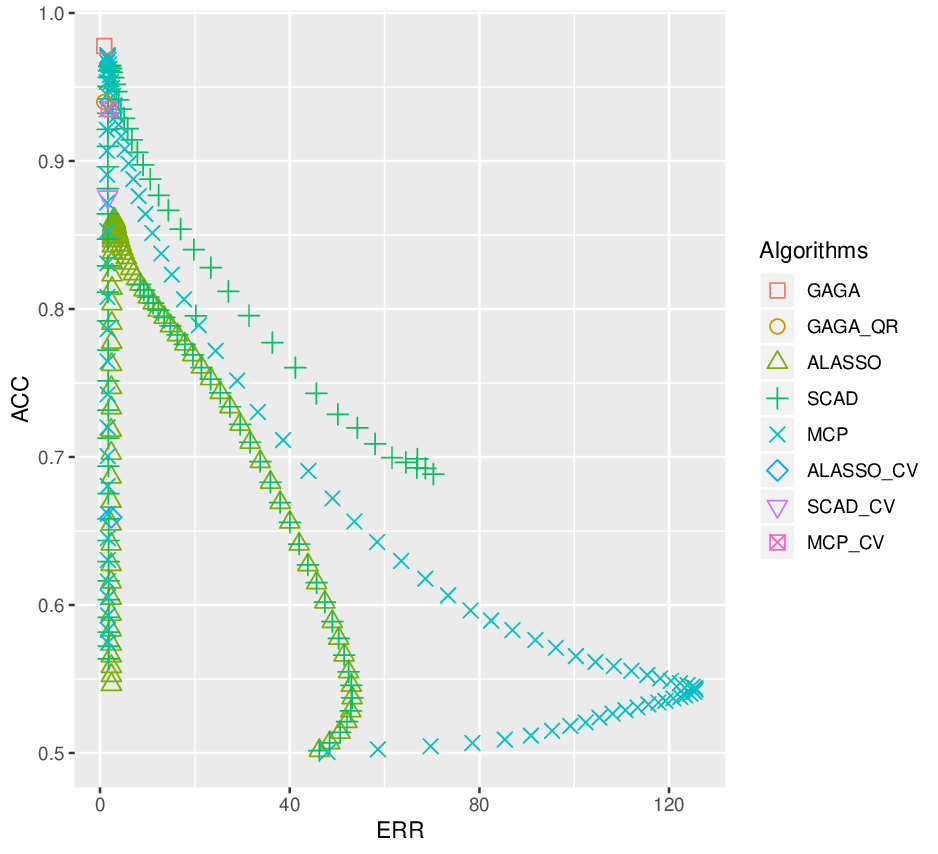}}
	\caption{Consider the ERR and the ACC comprehensively on the high-dimensional data.}\label{experiment3_ERR_ACC}
\end{figure}

For testing algorithms' performances on high-dimensional data, we simulate $100$ data sets from the linear model $y=\mathbf{x}^T\bm{\beta}^*+\varepsilon$. The true signal $\bm{\beta}^*$ has $500$ coefficients with $250$ zeros, whose positions are randomly generated in each data set.
The non-zero coefficient is randomly generated from $U(0,5)$. Each data set consists of $1000$ observations. For any $i\neq j$, $x_i$ and $x_j$ have a pairwise correlation of $0.5$.
The noise $\varepsilon$ is a standard normal random variable. Figures \ref{experiment3} and \ref{experiment3_ERR_ACC} illustrate that the GAGA also outperforms other penalized likelihood methods on the ERR and the ACC for the high-dimensional data.

\begin{figure}[h]
	\centering
	\subfigure[Performances on ERR] {\includegraphics[height=2.3in,width=2.7in]{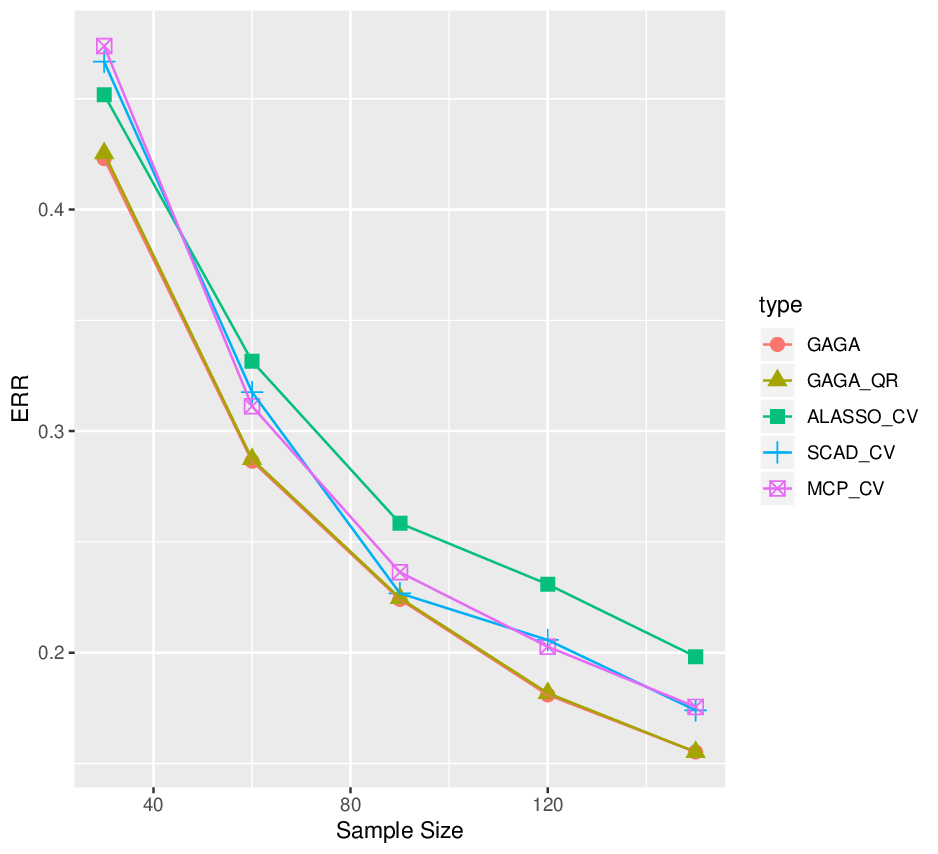}}
	\subfigure[Performances on ACC] {\includegraphics[height=2.3in,width=2.7in]{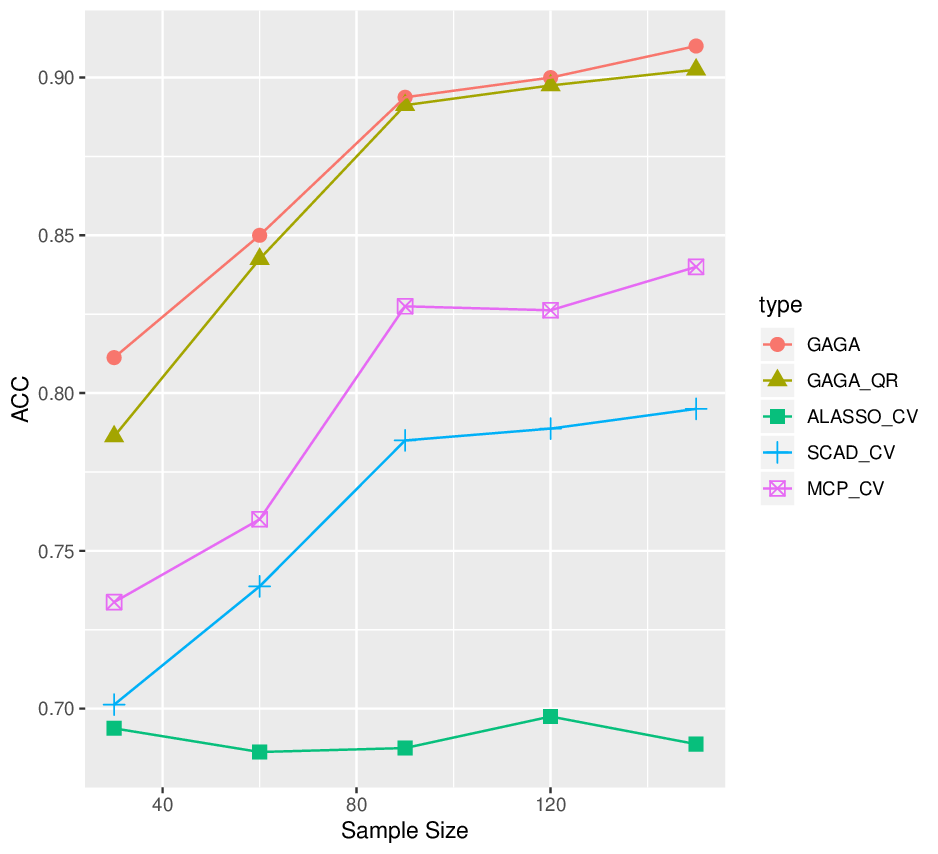}}
	\caption{The performance on the consistency of all the algorithms.}\label{experiment4}
\end{figure}

In the next experiment, we further consider the consistentcy of the output of algorithms as the sample size increases. The data is also generated from a linear model $y=\mathbf{x}^T\bm{\beta}^*+\varepsilon$, where the noise $\varepsilon$ is a standard normal random variable. The signal $\bm{\beta}^*$ has eight components. The number of non-zero coefficients in $\bm{\beta}$ is fixed to three, but the non-zero positions are random sampled. $100$ data sets are simulated on each sample size varying in $\{30,60,90,120,150\}$. We compute the average ERR and the average ACC for all the algorithms. For the adaptive LASSO, the SCAD and the MCP, we adopt the $10$-fold CV to select an appropriate tuning parameter for further signal estimates. The simulation result is shown in Figure \ref{experiment4}. As the sample size increases, the ERR of our algorithms goes down, and the ACC goes up. Moreover, the GAGA and the GAGA\_QR outperform other algorithms on the average ERR and the average ACC of $100$ data sets.

Finally, we compare the time costs between our algorithms and the adaptive LASSO, the SCAD, the MCP with $10$-fold Cross-Validation for the high-dimensional linear model. The dimension of the signal varies in $\{500,1000,2000\}$, the sample size is fixed to $4000$, and the experiment is repeated $10$ times for each dimension. Since the GAGA involves loops of the matrix's inversion, its time cost depends on the software's computational efficiency for numerical linear algebra. So we consider to execute the GAGA in both the Matlab and the R. The adaptive LASSO, the SCAD and the MCP are executed in the R with the ncvreg library \cite{ncvreg}. All the experiments in R adopt the OpenBlas for performing basic vector and matrix operations. As demenstrated in Table \ref{tab:1}, the GAGA\_QR in Matlab has the lowest average time costs on all the dimensions, and the SCAD has the highest ones. Even if considering the time costs only in R, the GAGA and the GAGA\_QR also outperform the other algorithms since those penalized likelihood methods have to go over all possible values of hyperparameter to pick up an appropriate one with the 10-fold CV. This experiment runs on a desktop with Intel i7 4.0 GHZ and 32 GB memory.

\begin{table}
	\caption{The average time cost of the GAGA, the GAGA\_QR, the adaptive LASSO, the SCAD and the MCP.}\label{tab:1}
	\centering
	\begin{tabular}{|c|c|c|c|c|c|c|c|}
		\hline
	Dimension	& $500$   &$1000$        & $2000$       \\ \hline
	GAGA(in Matlab)      & 0.36s &1.68s  &9.34s  \\\hline
	GAGA\_QR(in Matlab)	 & \textbf{0.15s} & \textbf{0.48s}& \textbf{1.74s} \\\hline
	GAGA(in R)  & 3.04s& 11.5s  & 66.67s \\\hline
	GAGA\_{QR}(in R)  & 1.45s& 5.09s  & 19.39s \\\hline
	ALASSO\_{CV}(in R)  & 27.36s& 54.12s  & 108.89s\\\hline
		SCAD\_{CV}(in R)  & 43.37s& 103.13s  & 221.37s \\\hline
			MCP\_{CV}(in R)  & 25.38s& 65.11s  & 152.51s \\\hline
\end{tabular}
\end{table}
\section{Conclusion}
In this paper, we propose an algorithm named GAGA for the signal recovery. This GAGA algorithm can automatically learn the hyperparamters, and update those hyperparameters and the signal in an alternative way. A variant algorithm called as GAGA\_QR is also suggested by using the QR-decompostion for improving the computational efficiency in the high dimensional analysis. In the theoretical part, we prove that the output of the GAGA algorithm can correctly find the positions of zero-coefficient with a high probability, and also provide a consistent estimate on the nonzero coefficients. 
In the simulation part, the experiment results illustrate that our algorithms outperform the adaptive Lasso, the SCAD and the MCP. Though the GAGA algorithm in this work is dedicated to the linear model, the mechanism behind the global adaptive generative adjustment strategy can be generalized to other statistical models. Exploration of this research direction is underway.

\bibliography{GAGA}
\bibliographystyle{icml2019}



\end{document}


\onecolumn
\icmltitle{Appendix to \\
	Global Adaptive Generative Adjustment}





\vskip 0.3in




\appendix
\section{Phase Change Phenomenon for Hyperparameters}
We denote $\mathcal{Q}$ as the subscript set $\{j|\beta_j^*\neq 0\}$ for non-zero components of signal. We assume that the design matrix $\mathbf{X}=(\textbf{x}_1,\cdots,\textbf{x}_p)$ is column orthogonal.
That is $\mathbf{X}^T\mathbf{X}=diag((a_1,\cdots,a_p))$.
And further assume that the condition number  $\kappa=\frac{\max_ja_j}{\min_ja_j}$ is bounded.
Since the column orthogonality of $\mathbf{X}$, the diagonal element
$(\mathbf{X}^T\mathbf{X}+\bm{\Lambda}^{(k)})^{-1}_{jj}=(a_j+\lambda_j^{(k)})^{-1}$.
So the update of $\lambda_j^{k+1}$ in Line 4 of the GAGA algorithm can be computed by $ \frac{\alpha(\lambda_j^{(k)}+a_j)^2}{\lambda_j^{(k)}+a_j+z_j}$ where $z_j=(\textbf{x}_j^T\textbf{y})^2$.
\begin{thm}\label{fixedpoint}
	For any $\alpha>1$ and any $j=1,\cdots,p$, we have the following conclusions:\\
	(1) when $z_j\geq\big((2\alpha-1)+2\sqrt{\alpha(\alpha-1)}\big)a_j$,
	the tuning parameter sequence $\{\lambda_j^{(k)}\}$ 
	converges to the fixed point $\lambda_j^*$ of $f_j(x)$
	if the sequence starts from an initial $\lambda_j^0=0$ and is generated by the update $\lambda_j^{k+1}=f_j(\lambda_j^{(k)})$, where $f_j(x) = \frac{\alpha(x+a_j)^2}{x+a_j+z_j},x\geq 0$.\\
	(2) when $ z_j< \big((2\alpha-1)+2\sqrt{\alpha(\alpha-1)}\big)a_j$,
	the tuning parameter sequence  $\{\lambda_j^{(k)}\}$ grows to the infinity.
\end{thm}

\begin{proof}
	Without loss of generalization, it suffices to consider the case $j=1$.
	Since $f_1'(x) = \frac{\alpha(x+a_1)^2+2\alpha z_1(a_1+x)}{(x+a_1+z_1)^2} >0$ for $x\geq 0$,
	we have that $f(x)$ is a strictly monotone increasing function. Moreover, $0<f_1'(x)<1$ when $0 \leq x < -a_1 + \big(\sqrt{\frac{\alpha}{\alpha-1}}-1\big)z_1$,
	and $f_1'(-a_1 + \big(\sqrt{\frac{\alpha}{\alpha-1}}-1\big)z_1)=1$.
	Since $\lambda_1^0=0$ and $\lambda_1^1=\frac{\alpha a_1^2}{a_1+z_1}>0$, we have that $\lambda_1^{k+1}-\lambda_1^{k}=f_1'(\xi_k)(\lambda_1^{(k)}-\lambda_1^{k-1})>0$ holds for any $k$. So $\{\lambda_1^{(k)}\}$ is a monotonic increasing sequence. Thus we have that $\lambda_1^{(k)}$ converges to a limit point which is a positive number or $\infty$. 
	
	$(1)$ Solve the equation $f_1(x) = x$,
	and we can obtain the fixed point
	\begin{equation}
	\lambda_1^* = \frac{z_1-(2\alpha-1)a_1-\sqrt{(z_1-(2\alpha-1)a_1)^2-4\alpha(\alpha-1)a_1^2}}{2(\alpha-1)}
	\end{equation}
	when $(z_1-(2\alpha-1)a_1)^2-4\alpha(\alpha-1)a_1^2\geq 0$. Notice that if $z_1\leq (2\alpha-1)a_1-2\sqrt{\alpha(\alpha-1)}a_1$, thus the fixed point $\lambda_1^*<0$. We only consider the case that $z_1\geq (2\alpha-1)a_1+ 2\sqrt{\alpha(\alpha-1)}a_1$. In this case, we have that $\lambda_1^*>0$ and 
	$\lambda_1^*-\lambda_1^{(k)}=f_1'(\xi_k)\cdots f_1'(\xi_1)\lambda_1^*>0$. Furthermore, we have that the sequence $\{\lambda_1^{(k)}\}$ is convergent since $\{\lambda_1^{(k)}\}$ is a monotonic increasing sequence.

	We first consider the case that $z_1= (2\alpha-1)a_1+ 2\sqrt{\alpha(\alpha-1)}a_1$. 
	We have that $\lambda_1^*=-a_1 + \big(\sqrt{\frac{\alpha}{\alpha-1}}-1\big)z_1=\sqrt{\frac{\alpha}{\alpha-1}}a_1$ when $z_1= (2\alpha-1)a_1+ 2\sqrt{\alpha(\alpha-1)}a_1$. Moreover,
	\begin{equation}\label{iter}
	\lambda_1^{k+1}-\lambda_1^{(k)} =  \frac{(\alpha-1)\bigg(\lambda_1^{(k)}-\sqrt{\frac{\alpha}{\alpha-1}}a_1\bigg)^2}{z_1+a_1+\lambda_1^{(k)}}.
	\end{equation}
	If $\lambda_1^{(k)}$ does not converge to $\lambda_1^*=\sqrt{\frac{\alpha}{\alpha-1}}a_1$, we have that $\lambda_1^{k+1}-\lambda_1^{(k)}>c$ for some constant $c>0$ when the iteration number $k$ is large enough. There is a contradiction with that $\{\lambda_1^{(k)}\}$ is convergent.
	
	Now we consider the case that $z_1> (2\alpha-1)a_1+ 2\sqrt{\alpha(\alpha-1)}a_1$.
	It can be verified that $\lambda_1^*<-a_1 + \big(\sqrt{\frac{\alpha}{\alpha-1}}-1\big)z_1$ in this case.
	Since $f''(x)=\frac{2\alpha z_1^2}{(x+a_1+z_1)^3}>0$ for $x\geq 0$ and $z_1> (2\alpha-1)a_1+ 2\sqrt{\alpha(\alpha-1)}a_1$, we know that $f'(x)$ is a strictly monotone increasing function. Thus $f'(b^*)<1$.
	Moreover, there are $f_1\big([0, \lambda_1^*]\big) \subseteq [0, \lambda_1^*]$ and $0<f_1'(x)\leq f_1'(\lambda_1^*)<1$ when $x\in[0,\lambda_1^*]$.
	By the fixed point iteration theorem,
	the iteration $ \lambda_1^{k+1} = \frac{\alpha(a_1+\lambda_1^{(k)})^2}{z_1+a_1+ \lambda_1^{(k)}}$ starting from $\lambda_1^0=0$ goes to the solution $\lambda_1^*$ of equation $x=f_1(x)$.
	
	Combining the above discussion, we get that the iteration $ \lambda_1^{k+1} = \frac{\alpha(a_1+\lambda_1^{(k)})^2}{z_1+a_1+ \lambda_1^{(k)}}$ starting from $\lambda_1^0=0$ goes to the solution $\lambda_1^*$ of equation $x=f_1(x)$ when $z_1\geq (2\alpha-1)a_1+ 2\sqrt{\alpha(\alpha-1)}a_1$.

	$(2)$ By the direct computation,
	\begin{equation}\label{iter}
	\begin{aligned}
	&\lambda_1^{k+1}-\lambda_1^{(k)} = \frac{(\alpha-1)(\lambda_1^{(k)})^2+\big((2\alpha-1)a_1-z_1\big)\lambda_1^{(k)}+\alpha a_1^2}{z_1+a_1+\lambda_1^{(k)}} \\
	&= \frac{(\alpha-1)\bigg(\lambda_1^{(k)}+\frac{(2\alpha-1)a_1-z_1}{2(\alpha-1)}\bigg)^2-\frac{\big((2\alpha-1)a_1-z_1\big)^2}{4(\alpha-1)}+\alpha a_1^2}{z_1+a_1+\lambda_1^{(k)}}.
	\end{aligned}
	\end{equation}
	
	If $~\alpha a_1^2-\frac{\big((2\alpha-1)a_1-z_1\big)^2}{4(\alpha-1)} > 0$,
	thus $~\big((2\alpha-1)-2\sqrt{\alpha(\alpha-1)}\big)a_1< z_1 <\big((2\alpha-1)+2\sqrt{\alpha(\alpha-1)}\big)a_1$
	($z_1>0$). Since $\lambda_1^{k}$ is an increasing sequence, we have $\lambda_1^{(k)}\rightarrow \infty$ as $k$ goes to infinity.
	(Otherwise, the bounded monotonic sequence $\{\lambda_1^{(k)}\}$ has a positve limit. This makes the difference $\lambda_1^{k+1}-\lambda_1^{(k)}$ is larger than a positive number $c$ when $k$ is large enough. It is a contradiction with that  $\{\lambda_1^{(k)}\}$ has a limit.)
	When $~z_1 \leq \big((2\alpha-1)-2\sqrt{\alpha(\alpha-1)}\big)a_1~$,
	we have that $\big((2\alpha-1)a_1-z_1\big)\geq2\sqrt{\alpha(\alpha-1)}a_1$.
	By similar discussion, we also obtain that $\lambda_1^{(k)}\rightarrow \infty$ as $k$ goes to infinity. So when $ z_1 < \big((2\alpha-1)+2\sqrt{\alpha(\alpha-1)}\big)a_1~$, we have that $\lambda_1^{k+1}-\lambda_1^{(k)}\asymp (\alpha-1)\lambda_1^{(k)}$ when $k$ is large enough. Thus $\lambda_1^{(k)}$ grows exponentially for large $k$.

\end{proof}

Furthermore, we show the probability of 
the event $\{z_j< \big((2\alpha-1)+2\sqrt{\alpha(\alpha-1)}\big)a_j\}$
depending on whether the true signal $\beta_j^*=0$ or not. Let $g$ be a standard normal random variable.
\begin{lem}\label{zscope}
	 For $\alpha>1$, we have that\\
	(1) If the true signal $\beta_j^*\neq 0$, $P(z_j< \big((2\alpha-1)+2\sqrt{\alpha(\alpha-1)}\big)a_j)\leq 2\exp(-\frac{1}{2}(\sqrt{a_j}|\beta_j^*|-(\sqrt{\alpha}+\sqrt{\alpha-1}))^2)$ 
	when the inequality $\sqrt{a_j}|\beta_j|-(\sqrt{\alpha}+\sqrt{\alpha-1})\geq 0$ holds.\\
	(2) If the true signal $\beta_j^*= 0$, $P(z_j< \big((2\alpha-1)+2\sqrt{\alpha(\alpha-1)}\big)a_j)\geq 1-2\exp(-\frac{1}{2}((2\alpha-1)+\sqrt{\alpha(\alpha-1)})).$ 
\end{lem}

\begin{proof}
	We know that $\mathbf{x}_j^T\mathbf{y}=a_j\beta_j^*+\sqrt{a_j}g$ where $g$ be a standard normal random variable.
	When the true signal $~\beta_j^* \neq 0~$, we have that
	\begin{equation*}
	\begin{aligned}
	 \mathbb{P}\Big(z_j<\big((2\alpha-1)+2\sqrt{\alpha(\alpha-1)}\big)a_j\Big)
	=& \mathbb{P}\Big((\mathbf{x}_j^T\textbf{y})^2<\big((2\alpha-1)+2\sqrt{\alpha(\alpha-1)}\big)a_j\Big)\\
	\leq & \mathbb{P}(|g|>\sqrt{a_j}|\beta_j|-(\sqrt{\alpha}+\sqrt{\alpha-1}))\\
	\leq &\exp(-\frac{1}{2}(\sqrt{a_j}|\beta_j|-(\sqrt{\alpha}+\sqrt{\alpha-1}))^2).
	\end{aligned}
	\end{equation*}
	When $~\beta_j^* = 0~$, we have that
	\begin{equation*}
	\begin{aligned}
	\mathbb{P}\Big(z_j< \big((2\alpha-1)+2\sqrt{\alpha(\alpha-1)}\big)a_j\Big)
	=& \mathbb{P}(|g|<\sqrt{\alpha}+\sqrt{\alpha-1})\\
	\geq & 1-\exp(-\frac{1}{2}((2\alpha-1)+\sqrt{\alpha(\alpha-1)})).
	\end{aligned}
	\end{equation*}
\end{proof}

\begin{lem}\label{En}
	 Denote $\mathcal{Q}$ as the subscript set $\{j|\beta_j^*\neq 0\}$ for non-zero components of signal. We have that
	$$\mathbb{P}(E)	\geq
	1-\exp(-\frac{1}{2}(\sqrt{\alpha}+\sqrt{\alpha-1})^2+\log(p-q))$$
	where the event $E=\{z_j <(\sqrt{\alpha}+\sqrt{\alpha-1})^2a_j,\forall j\in\mathcal{Q}^c\}$.
\end{lem}

\section{Properties of the Hyperparameter Limit}

\begin{thm}\label{fixed}
	For the true signal $\beta_j^*\neq 0$ and the growth rate $\alpha>1$, let $x_j^*$ be the fixed point $\frac{z_j-(2\alpha-1)a_j-\sqrt{((2\alpha-1)a_j-z_j)^2-4\alpha(\alpha-1)a_j^2}}{2(\alpha-1)}$ of the update function $f_j(x) = \frac{\alpha(x+a_j)^2}{x+a_j+z_j}$.
	We have that for any $0<\eta<1$,
	$$\mathbb{P}(\frac{\alpha}{x_j^*}<\eta^2{\beta_j^*}^2,z_j>(\sqrt{\alpha}+\sqrt{\alpha-1})^2a_j)\leq \exp(-\frac{1}{2}((1-\eta)a_j^{1/2}|\beta_j^*|-(\sqrt{\alpha}+\sqrt{\alpha-1}))^2)$$
	when $(1-\eta)a_j^{1/2}|\beta_j^*|-(\sqrt{\alpha}+\sqrt{\alpha-1})\geq 0$.
\end{thm}

\begin{proof}
	If $z_j>((2\alpha-1)+2\sqrt{\alpha(\alpha-1)})a_j$,
	we can get
	\begin{equation*}
	\begin{aligned}
    &((z_j-(2\alpha-1)a_j)^2)^{1/2}-((z_j-(2\alpha-1)a_j)^2-4\alpha(\alpha-1)a_j^2)^{1/2}\\
	\leq & 4\alpha(\alpha-1)a_j^2/(2((z_j-(2\alpha-1)a_j)^2-4\alpha(\alpha-1)a_j^2)^{1/2})
	\end{aligned}
	\end{equation*}
	by mean value theorem.
	Furthermore, we have that $x_j^*/\alpha\leq a_j^2/((z_j-(2\alpha-1)a_j)^2-4\alpha(\alpha-1)a_j^2)^{1/2}$
	if $z_j>((2\alpha-1)+2\sqrt{\alpha(\alpha-1)})a_j$.
	Let us further consider the probability
	$$p_*=\mathbb{P}(\frac{((z_j-(2\alpha-1)a_j)^2-4\alpha(\alpha-1)a_j^2)^{1/2}}{a_j^2}<\eta^2{\beta_j^*}^2,z_j>((2\alpha-1)+2\sqrt{\alpha(\alpha-1)})a_j).$$
	It equals to the probability 
	$$\mathbb{P}((\sqrt{\alpha}+\sqrt{\alpha-1})a_j^{1/2}\leq|\mathbf{x}_j^T\mathbf{y}|<((2\alpha-1)a_j+(4\alpha(\alpha-1)a_j^2+a_j^4\eta^4{\beta_j^*}^4)^{1/2})^{1/2}).$$
	We know that $\mathbf{x}_j^T\mathbf{y}=a_j\beta_j^*+\sqrt{a_j}g$ where $g$ be a standard normal random variable.
	We have that
	\begin{equation*}
		\begin{aligned}
			p_*\leq& \mathbb{P}(a_j^{1/2}|g|>a_j|\beta_j^*|-((2\alpha-1)a_j+(4\alpha(\alpha-1)a_j^2+a_j^4\eta^4{\beta_j^*}^4)^{1/2})^{1/2})\\
			\leq & \mathbb{P}(a_j^{1/2}|g|>(1-\eta)a_j|\beta_j^*|-(\sqrt{\alpha}+\sqrt{\alpha-1})a_j^{1/2})\\
			= & \mathbb{P}(|g|>(1-\eta)a_j^{1/2}|\beta_j^*|-(\sqrt{\alpha}+\sqrt{\alpha-1}))\\
			\leq & \exp(-\frac{1}{2}((1-\eta)a_j^{1/2}|\beta_j^*|-(\sqrt{\alpha}+\sqrt{\alpha-1}))^2)
		\end{aligned}
	\end{equation*}
		when $(1-\eta)a_j^{1/2}|\beta_j^*|-(\sqrt{\alpha}+\sqrt{\alpha-1})\geq 0$.
\end{proof}

For any convergent subsequence $\{\bm{\lambda}^{k_l}\}$ of $\{\bm{\lambda}^{k}\}$,
denote its limit $\lim\limits_{l\rightarrow\infty}\bm{\lambda}^{k_l}$ by $\bm{\lambda}^\infty$. 
The following theorem shows $\min\limits_{j\in \mathcal{Q}}\frac{\alpha}{\lambda_j^\infty{\beta_j^*}^2}$ has a lower bound with a high probability. 
\begin{thm}\label{blimit}
	Suppose that Assumptions (A1) to (A3) hold. For any $0<\eta<1$, 
	we have that
	\begin{equation*}
	\begin{aligned}
		\mathbb{P}(\bigcup\limits_{j\in\mathcal{Q}}\{\frac{\alpha}{\lambda_j^\infty{\beta_j^*}^2}<\eta^2\})\leq& \exp(-\frac{1}{2}((1-\eta)\min\limits_{j\in \mathcal{Q}}a_j^{1/2}|\beta_j^*|-(\sqrt{\alpha}+\sqrt{\alpha-1}))^2+\log(q))\\
		&+\exp(-\frac{1}{2}(\min\limits_{j\in \mathcal{Q}}a_j^{1/2}|\beta_j|-(\sqrt{\alpha}+\sqrt{\alpha-1}))^2+\log(q)).
	\end{aligned}
\end{equation*}
when $(1-\eta)\min\limits_{j\in \mathcal{Q}}a_j^{1/2}|\beta_j^*|-(\sqrt{\alpha}+\sqrt{\alpha-1})\geq 0$.
\end{thm}

\begin{proof}
	It suffices to consider the lower bound of the probability
	$\mathbb{P}(\bigcap\limits_{j\in\mathcal{Q}}\{\frac{\alpha}{\lambda_j^\infty{\beta_j^*}^2}\geq \eta^2\})$.
	We only need to show the upper bound of the probability $\mathbb{P}(\bigcup\limits_{j\in\mathcal{Q}}\{\frac{\alpha}{\lambda_j^\infty{\beta_j^*}^2}<\eta^2\})$.
	Notice that 
	$$f_j(\lambda_j^{(k)})=\lambda_j^{(k+1)}=\frac{\alpha}{\frac{z_j}{(a_j+\lambda_j^{(k)})^2}+\frac{1}{a_j+\lambda_j^{(k)}}}.$$
	From Theorem \ref{fixedpoint} (1), we know that the limit $\lambda_j^\infty$ is the fixed point of the update function $f_j(x)$ in case that
	$z_j\geq(\sqrt{\alpha}+\sqrt{\alpha-1}\big)^2a_j$. Let the event  $G=\bigcap\limits_{j\in\mathcal{Q}}\{L_j\geq(\sqrt{\alpha}+\sqrt{\alpha-1}\big)\sqrt{a_{11}{(j)}}\}.$
	Thus 
	\begin{equation*}
		\begin{aligned}
			\mathbb{P}(\bigcup\limits_{j\in\mathcal{Q}}\{\frac{\alpha}{\lambda_j^\infty{\beta_j^*}^2}<\eta^2\},G)
			= & \mathbb{P}(\bigcup\limits_{j\in\mathcal{Q}}\{\frac{\alpha}{\lambda_j^\infty}<\eta^2{\beta_j^*}^2\},G)\\
			\leq &\sum\limits_{j\in\mathcal{Q}}\mathbb{P}(\frac{\alpha}{\lambda_j^\infty}<\eta^2{\beta_j^*}^2,L_j\geq(\sqrt{\alpha}+\sqrt{\alpha-1}\big)\sqrt{a_{11}{(j)}})\\
			\leq &\sum\limits_{j\in\mathcal{Q}}\exp(-\frac{1}{2}((1-\eta)a_j^{1/2}|\beta_j^*|-(\sqrt{\alpha}+\sqrt{\alpha-1}))^2)\\
			\leq& \exp(-\frac{1}{2}((1-\eta)\min\limits_{j\in \mathcal{Q}}a_j^{1/2}|\beta_j^*|-(\sqrt{\alpha}+\sqrt{\alpha-1}))^2+\log(q))
		\end{aligned}
	\end{equation*}
	by Theorem \ref{fixed}.
	Furthermore, 
	we also have that
	$$\mathbb{P}(G^c)\\
	\leq\exp(-\frac{1}{2}(\min\limits_{j\in \mathcal{Q}}a_j^{1/2}|\beta_j|-(\sqrt{\alpha}+\sqrt{\alpha-1}))^2+\log(q))$$
	by Lemma \ref{zscope}. So we have that
	\begin{equation*}
		\begin{aligned}
			\mathbb{P}(\bigcup\limits_{j\in\mathcal{Q}}\{\frac{\alpha}{\lambda_j^\infty{\beta_j^*}^2}<\eta^2\})\leq& \exp(-\frac{1}{2}((1-\eta)\min\limits_{j\in \mathcal{Q}}a_j^{1/2}|\beta_j^*|-(\sqrt{\alpha}+\sqrt{\alpha-1}))^2+\log(q))\\
			&+\exp(-\frac{1}{2}(\min\limits_{j\in \mathcal{Q}}a_j^{1/2}|\beta_j|-(\sqrt{\alpha}+\sqrt{\alpha-1}))^2+\log(q)).
		\end{aligned}
	\end{equation*}
when $(1-\eta)\min\limits_{j\in \mathcal{Q}}a_j^{1/2}|\beta_j^*|-(\sqrt{\alpha}+\sqrt{\alpha-1})\geq 0$.
\end{proof}

For any convergent subsequence $\{\bm{\lambda}^{k_l}\}_l$ of $\{\bm{\lambda}^{k}\}_k$,
consider its limit $\bm{\lambda}^\infty=\lim\limits_{l\rightarrow\infty}\bm{\lambda}^{k_l}$ and $\bm{\lambda}^*=\bm{\lambda}^{\infty}/\alpha$.
Theorem \ref{blimit} shows that
\begin{equation*}
	\begin{aligned} \mathbb{P}(\max\limits_{j\in\mathcal{Q}}\lambda_j^*{\beta_j^*}^2\leq \frac{1}{\eta^2})=&\mathbb{P}(\max\limits_{j\in\mathcal{Q}}\frac{\lambda_j^{\infty}{\beta_j^*}^2}{\alpha}\leq\frac{1}{\eta^2})\\
		=&\mathbb{P}(\min\limits_{j\in \mathcal{Q}}\frac{\alpha}{\lambda_j^{\infty}{\beta_j^*}^2}\geq \eta^2)\\
		\geq& 1-\exp(-\frac{1}{2}((1-\eta)\min\limits_{j\in \mathcal{Q}}a_j^{1/2}|\beta_j^*|-(\sqrt{\alpha}+\sqrt{\alpha-1}))^2+\log(q))\\
		&-\exp(-\frac{1}{2}(\min\limits_{j\in \mathcal{Q}}a_j^{1/2}|\beta_j|-(\sqrt{\alpha}+\sqrt{\alpha-1}))^2+\log(q)).
	\end{aligned}
\end{equation*}
for any $0<\eta<1$. 
\begin{lem}\label{Fevent}
	Denote $F$ as the event $\{\max\limits_{j\in\mathcal{Q}}\lambda_j^*{\beta_j^*}^2\leq\frac{1}{\eta^2}\}$
	where $0<\eta<1$.
	We have that 
	\begin{equation*}
		\begin{aligned}
			\mathbb{P}(F^c)\leq& \exp(-\frac{1}{2}((1-\eta)\min\limits_{j\in \mathcal{Q}}a_j^{1/2}|\beta_j^*|-(\sqrt{\alpha}+\sqrt{\alpha-1}))^2+\log(q))\\
			&+\exp(-\frac{1}{2}(\min\limits_{j\in \mathcal{Q}}a_j^{1/2}|\beta_j|-(\sqrt{\alpha}+\sqrt{\alpha-1}))^2+\log(q))
		\end{aligned}
	\end{equation*}
when $(1-\eta)\min\limits_{j\in \mathcal{Q}}a_j^{1/2}|\beta_j^*|-(\sqrt{\alpha}+\sqrt{\alpha-1})\geq 0$.
\end{lem}

\section{Proof of Theorem 2.1}
\begin{proof}
	For any convergent subsequence $\{\bm{\lambda}^{k_l}\}_l$ of $\{\bm{\lambda}^{k}\}_k$,
	consider its limit $\bm{\lambda}^\infty=\lim\limits_{l\rightarrow\infty}\bm{\lambda}^{k_l}$ and $\bm{\lambda}^*=\bm{\lambda}^{\infty}/\alpha$. Let $\hat{\bm{\beta}}=(\mathbf{X}^T\mathbf{X}+\bm{\Lambda}^*)^{-1}\mathbf{X}^T\mathbf{y}$ where $\bm{B}^*=diag(\bm{\lambda}^*)$.
	Note that if the event 
	$$E=\{z_j <(\sqrt{\alpha}+\sqrt{\alpha-1})^2a_j,\forall j\in\mathcal{Q}^c\}$$ happens, we have that $\lambda_j^\infty=\infty$ for any $j\in\mathcal{Q}^c$ by Theorem \ref{fixedpoint}. So $\hat{\beta}_j=0$.
	Furthermore, 
	by Lemma \ref{En},
	\begin{equation*}
		\begin{aligned}
			&\mathbb{P}(\bigcap\limits_{j\in\mathcal{Q}^c}\{(\hat{\beta}_j)^2\leq a_j^{-1}-(a_j+\lambda_j^*)^{-1})_{jj}\})\\
			\geq&\mathbb{P}(\bigcap\limits_{j\in \mathcal{Q}^c}\{(\hat{\beta}_j)^2=0\})\\
			\geq& \mathbb{P}(\bigcap\limits_{j\in \mathcal{Q}^c}\{(\hat{\beta}_j)^2=0\},E)=\mathbb{P}(E)\\	
			\geq
			&1-\exp(-\frac{1}{2}(\sqrt{\alpha}+\sqrt{\alpha-1})^2+\log(p-q)).\\
		\end{aligned}
	\end{equation*}

	Recall that $F=\{\max\limits_{j\in\mathcal{Q}}\lambda_j^*{\beta_j^*}^2\leq\frac{1}{\eta^2}\}$ where $0<\eta<1$.
	We consider the probability of the personalized thresholding event:
	\begin{equation*}
		\begin{aligned}
			&\mathbb{P}(\bigcup\limits_{j\in\mathcal{Q}}\{|\hat{\beta}_j|^2\leq a_j^{-1}-(a_j+\lambda_j^*)^{-1}\})\\
			\leq & \mathbb{P}(\bigcup\limits_{j\in\mathcal{Q}}\{|\hat{\beta}_j|\leq(a_j^{-1}-(a_j+\lambda_j^*)^{-1})^{1/2}\},F)+\mathbb{P}(F^c).
		\end{aligned}
	\end{equation*}
	By Lemma \ref{Fevent}, the second term
	\begin{equation*}
		\begin{aligned}
			\mathbb{P}(F^c)\leq& \exp(-\frac{1}{2}((1-\eta)\min\limits_{j\in \mathcal{Q}}a_j^{1/2}|\beta_j^*|-(\sqrt{\alpha}+\sqrt{\alpha-1}))^2+\log(q))\\
			&+\exp(-\frac{1}{2}(\min\limits_{j\in \mathcal{Q}}a_j^{1/2}|\beta_j|-(\sqrt{\alpha}+\sqrt{\alpha-1}))^2+\log(q))
		\end{aligned}
	\end{equation*}
when $(1-\eta)\min\limits_{j\in \mathcal{Q}}a_j^{1/2}|\beta_j^*|-(\sqrt{\alpha}+\sqrt{\alpha-1})\geq 0$.
	It suffices to study the first term. 
	Notice that $(\hat{\beta}_j-\frac{a_j\beta_j}{a_j+\lambda_j^*})*(a_j+\lambda_j^*)\sim N(0,a_j)$
	for any $j\in\mathcal{Q}$. 
	Let $g$ be a standard normal random variable.
	Furthermore, 
	\begin{equation*}
		\begin{aligned}
			&\mathbb{P}(\bigcup\limits_{j\in\mathcal{Q}}\{|\hat{\beta}_j|\leq(a_j^{-1}-(a_j+\lambda_j^*)^{-1})^{1/2}\},F)\\
			\leq &\mathbb{P}(\bigcup\limits_{j\in\mathcal{Q}}\{|g|\geq a_j^{1/2}|\beta_j^*|-\frac{{\lambda_j^*}^{1/2}(a_j+\lambda_j^*)^{1/2}}{a_{j}}\},F)\\
			\leq
			&\mathbb{P}(\bigcup\limits_{j\in\mathcal{Q}}\{|g|\geq a_j^{1/2}|\beta_j^*|-(1+\frac{\lambda_j^*}{a_j})\},F).\\
		\end{aligned}
	\end{equation*}

	Notice that
	$$\frac{\lambda_j^*}{a_j}\leq  \frac{\max\limits_{j\in\mathcal{Q}}\lambda_j^*{\beta_j^*}^2}{\min\limits_{j\in \mathcal{Q}}a_j{\beta_j^*}^2}\leq\frac{1}{\eta^2\min\limits_{j\in \mathcal{Q}}a_j{\beta_j^*}^2}$$
	when the event $F$ happens. 
	So
	\begin{equation*}
		\begin{aligned}
			&\mathbb{P}(\bigcup\limits_{j\in\mathcal{Q}}\{|\hat{\beta}_j|\leq(a_j^{-1}-(a_j+\lambda_j^*)^{-1})^{1/2}\},F)\\
			\le&\mathbb{P}(\bigcup\limits_{j\in\mathcal{Q}}\{|g|\geq a_j^{1/2}|\beta_j^*|-(1+\frac{\lambda_j^*}{a_j})\},F)\\
			\leq&\mathbb{P}(\bigcup\limits_{j\in\mathcal{Q}}\{|g|\geq a_j^{1/2}|\beta_j^*|-(1+\frac{1}{\eta^2\min\limits_{j\in \mathcal{Q}}a_j{\beta_j^*}^2}),F)\\
			\leq&\exp(-\frac{1}{2}(\min\limits_{j\in \mathcal{Q}}a_j^{1/2}|\beta_j^*|-(1+\frac{1}{\eta^2\min\limits_{j\in \mathcal{Q}}a_j{\beta_j^*}^2}))^2+\log(q))\\
			\leq&\exp(-\frac{1}{2}(\min\limits_{j\in \mathcal{Q}}a_j^{1/2}|\beta_j^*|-(\sqrt{\alpha}+\sqrt{\alpha-1}))^2+\log(q))
		\end{aligned}
	\end{equation*}
	when $\eta^2\min\limits_{j\in \mathcal{Q}}a_j{\beta_j^*}^2\geq \frac{1}{\sqrt{\alpha}+\sqrt{\alpha-1}-1}$ and $(1-\eta)\min\limits_{j\in \mathcal{Q}}a_j^{1/2}|\beta_j^*|-(\sqrt{\alpha}+\sqrt{\alpha-1})\geq 0$. Thus we complete the proof. 	
\end{proof}

\section{Proof of Theorem 2.2}

Let the true signal $\bm{\beta}^*=({\bm{\beta}_1^*}^T,{\bm{\beta}_2^*}^T)^T$. Without loss of generalization, assume that $\bm{\beta}_1^*$ consists of $q$ non-zero components and $\bm{\beta}_2^*$ consists of all zero components. So $\mathcal{Q}=\{1,\cdots,q\}$. 
According to the true signal $\bm{\beta}^*$, the hyperparameter vector $\bm{\lambda}$ can be also divided into $\bm{\lambda}_1$ and $\bm{\lambda}_2$ for penalizing $\bm{\beta}_1^*$ and $\bm{\beta}_2^*$ respectively.
Consider the hyperparameter limit $\bm{\lambda}^\infty=\lim\limits_{l\rightarrow\infty}\bm{\lambda}^{k_l}$ where $\{\bm{\lambda}^{k_l}\}_l$ is any convergent subsequence of $\{\bm{\lambda}^{k}\}_k$. 
Let $\bm{\Lambda}^*=diag(\bm{\lambda}^{\infty})/\alpha$, $\hat{\bm{\beta}}=(\mathbf{X}^T\mathbf{X}+\bm{\Lambda}^*)^{-1}\mathbf{X}^T\mathbf{y}$ and $\kappa=\frac{\max_ja_j}{\min_ja_j}$.
\begin{lem}\label{errorbound}
	Let the events 	$E=\{z_j <(\sqrt{\alpha}+\sqrt{\alpha-1})^2a_j,j\in\mathcal{Q}^c\}$
	and 
	$F=\{\max\limits_{j\in\mathcal{Q}}\lambda_j^*{\beta_j^*}^2\leq\frac{1}{\eta^2}\}$ where $0<\eta<1$,
	we have that 
	\begin{equation*}
		\begin{aligned}
			E\|\hat{\bm{\beta}}-\bm{\beta}^*\|\leq & \frac{\sqrt{q}}{\min\limits_{j\in \mathcal{Q}}a_j^{1/2}}(\sqrt{\kappa}+\frac{1}{\eta^2\min\limits_{j\in \mathcal{Q}}a_j^{1/2}|\beta_j^*|})+\\
			&(E\|\hat{\bm{\beta}}\|^2)^{1/2}((\mathbb{P}(E^c))^{1/2}+(\mathbb{P}(F^c))^{1/2})+  \|\bm{\beta}^*\|(\mathbb{P}(E^c)+\mathbb{P}(F^c)).\\
		\end{aligned}
	\end{equation*}
\end{lem}
\begin{proof}
	We consider the expectation
	$$\mathbb{E}\|\hat{\bm{\beta}}-\bm{\beta}^*\|=\mathbb{E}\|\hat{\bm{\beta}}-\bm{\beta}^*\|I_{E}+\mathbb{E}\|\hat{\bm{\beta}}-\bm{\beta}^*\|I_{E^c}.$$
	where $\hat{\bm{\beta}}=(\mathbf{X}^T\mathbf{X}+\bm{\Lambda^*})^{-1}\mathbf{X}^T\mathbf{y}$.
	For the second term, we have that 
	\begin{equation*}
		\begin{aligned}
			\mathbb{E}\|\hat{\bm{\beta}}-\bm{\beta}^*\|I_{E^c}\leq & \mathbb{E}\|\hat{\bm{\beta}}\|I_{E^c}+\mathbb{E}\|\bm{\beta}^*\|I_{E^c}\\
			\leq & (\mathbb{E}\|\hat{\bm{\beta}}\|^2)^{1/2}(\mathbb{P}(E^c))^{1/2} + \|\bm{\beta}^*\|\mathbb{P}(E^c).\\
		\end{aligned}
	\end{equation*}
	For the first term $\mathbb{E}\|\hat{\bm{\beta}}-\bm{\beta}^*\|I_{E}$,we know that 
	$$\mathbb{E}\|\hat{\bm{\beta}}-\bm{\beta}^*\|I_{E}= \mathbb{E}\|\hat{\bm{\beta}}-\bm{\beta}^*\|I_{E}I_{F}+\mathbb{E}\|\hat{\bm{\beta}}-\bm{\beta}^*\|I_{E}I_{F^c}$$
	where  $F$ is the event $\{\max\limits_{j\in\mathcal{Q}}\lambda_j^*{\beta_j^*}^2\leq\frac{1}{\eta^2}\}$.
	Further, we have that 
	\begin{equation*}
		\begin{aligned}
			\mathbb{E}\|\hat{\bm{\beta}}-\bm{\beta}^*\|I_{E}I_{F^c}\leq & \mathbb{E}\|\hat{\bm{\beta}}\|I_{F^c}+\mathbb{E}\|\bm{\beta}^*\|I_{F^c}\\
			\leq & (\mathbb{E}\|\hat{\bm{\beta}}\|^2)^{1/2}(\mathbb{P}(F^c))^{1/2} + \|\bm{\beta}^*\|\mathbb{P}(F^c).\\
		\end{aligned}
	\end{equation*}
	When $E$ holds,
	we have that $\lambda_j^*=\infty$ for any $j\in\mathcal{Q}^c$.  Futhermore, $\hat{\bm\beta}_2=\mathbf{0}$ and by the column orthogonality
	\begin{equation*}
		\begin{aligned}
			\hat{\bm{\beta}}_1
			=&(diag(a_1+\lambda_1,\cdots,a_q+\lambda_q))^{-1}\mathbf{X}_1^T\mathbf{y}\\
			=&diag((\frac{a_1}{a_1+\lambda_1},\cdots,\frac{a_q}{a_q+\lambda_q}))\bm{\beta}_1^*+(diag(a_1+\lambda_1,\cdots,a_q+\lambda_q))^{-1}\mathbf{X}_1^T\bm{\epsilon}
		\end{aligned}
	\end{equation*}
	And further we have that
	\begin{equation*}
		\begin{aligned}
			\mathbb{E}\|\hat{\bm{\beta}}-\bm{\beta}^*\|I_{E}I_{F}\leq &	\mathbb{E}\|\hat{\bm{\beta}}_1-\bm{\beta}_1^*\|I_{F}\\
			\leq & \mathbb{E}\|diag((\frac{\lambda_1}{a_1+\lambda_1},\cdots,\frac{\lambda_q}{a_q+\lambda_q}))\bm{\beta}_1^*\|I_{F}+\\
			&\mathbb{E}\|(diag(a_1+\lambda_1,\cdots,a_q+\lambda_q))^{-1}\mathbf{X}_1^T\bm{\epsilon}\|I_{F}\\		
		\end{aligned}
	\end{equation*}
	Since 		$\mathbb{E}\|\mathbf{X}_1^T\bm{\epsilon}\|\leq(\mathbb{E}\|\mathbf{X}_1^T\bm{\epsilon}\|^2)^{1/2}\leq (\sum\limits_{j\in \mathcal{Q}}a_j)^{1/2},$
	thus $$\mathbb{E}\|(diag(a_1+\lambda_1,\cdots,a_q+\lambda_q))^{-1}\mathbf{X}_1^T\bm{\epsilon}\|I_{F}\leq\frac{ (\sum\limits_{j\in \mathcal{Q}}a_j)^{1/2}}{\min\limits_{j\in \mathcal{Q}}a_j}.$$ 
	And by the definition of the set $F$,
	$$\mathbb{E}\|diag((\frac{\lambda_1}{a_1+\lambda_1},\cdots,\frac{\lambda_q}{a_q+\lambda_q}))\bm{\beta}_1^*\|I_{F}\leq \mathbb{E}\frac{\sqrt{q}\max\limits_{j\in\mathcal{Q}}\lambda_j^*|\beta_j^*|^2}{\min\limits_{j\in \mathcal{Q}}a_j|\beta_j^*|}I_{F}\leq\frac{\sqrt{q}}{\eta^2\min\limits_{j\in \mathcal{Q}}a_j|\beta_j^*|}.$$
	So 
	\begin{equation*}
		\begin{aligned}
			\mathbb{E}\|\hat{\bm{\beta}}_1-\bm{\beta}_1^*\|I_{F}\leq \frac{(\sum\limits_{j\in \mathcal{Q}}a_j)^{1/2}}{\min\limits_{j\in \mathcal{Q}}a_j}+\frac{\sqrt{q}}{\eta^2\min\limits_{j\in \mathcal{Q}}a_j|\beta_j^*|}\leq
			\frac{\sqrt{q}}{\min\limits_{j\in \mathcal{Q}}a_j^{1/2}}(\sqrt{\kappa}+\frac{1}{\eta^2\min\limits_{j\in \mathcal{Q}}a_j^{1/2}|\beta_j^*|}).\\		
		\end{aligned}
	\end{equation*}
	Thus we get the conclusion.
\end{proof}

\begin{proof}
	
	Now we consider the difference between $\hat{\bm{\beta}}$ and the personalized thresholding $\hat{\bm{\beta}}^*$. $$E\|\hat{\bm{\beta}}^*-\hat{\bm{\beta}}\|=E\|\hat{\bm{\beta}}^*-\hat{\bm{\beta}}\|I_{E}+E\|\hat{\bm{\beta}}^*-\hat{\bm{\beta}}\|I_{E^c}.$$
	We have that $E\|\hat{\bm{\beta}}^*-\hat{\bm{\beta}}\|I_{E^c}\leq E\|\hat{\bm{\beta}}\|I_{E^c}\leq (E\|\hat{\bm\beta}\|^2)^{1/2}(\mathbb{P}(E^c))^{1/2}$, and
	$E\|\hat{\bm{\beta}}^*-\hat{\bm{\beta}}\|I_{E}=E\|\hat{\bm{\beta}}_1^*-\hat{\bm{\beta}}_1\|I_{E}$.
	Since
	$$\|\hat{\bm{\beta}}_1^*-\hat{\bm{\beta}}_1\|^2\leq\sum\limits_{j\in \mathcal{Q},\hat{\beta}_j \text{is truncated.}}|\hat{\beta}_j|^2\leq\sum\limits_{j\in \mathcal{Q}} a_j^{-1}\leq\frac{q}{\min\limits_{j\in \mathcal{Q}}a_j}$$
	by the truncation condition,
	thus $E\|\hat{\bm{\beta}}^*-\hat{\bm{\beta}}\|I_{E}=E\|\hat{\bm{\beta}}_1^*-\hat{\bm{\beta}}_1\|I_{E}\leq \sqrt{\frac{q}{\min\limits_{j\in \mathcal{Q}}a_j}}$. Furthermore, using Lemma \ref{errorbound}
	we have that 
	\begin{equation*}
		\begin{aligned}
			&\mathbf{E}\|\hat{\bm{\beta}}^*-\bm{\beta}^*\|\\
			\leq & \sqrt{\frac{q}{\min\limits_{j\in \mathcal{Q}}a_j}}(1+\sqrt{\kappa}+\frac{1}{\eta^2\min\limits_{j\in \mathcal{Q}}a_j^{1/2}|\beta_j^*|})+\\
			& (E\|\hat{\bm{\beta}}\|^2)^{1/2}(2(\mathbb{P}(E^c))^{1/2}+(\mathbb{P}(F^c))^{1/2})
			+  \|\bm{\beta}^*\|(\mathbb{P}(E^c)+\mathbb{P}(F^c)).\\
		\end{aligned}
	\end{equation*}
	In the following part, we further discuss the upper bound of $\mathbf{E}\|\hat{\bm{\beta}}^*-\bm{\beta}^*\|$.
	Since $\|\hat{\bm{\beta}}\|^2=\mathbf{y}^T\mathbf{X}(\mathbf{X}^T\mathbf{X}+\bm{\Lambda})^{-2}\mathbf{X}^T\mathbf{y}$ and $\mathbf{X}^T\mathbf{X}+\bm{\Lambda}\geq\min\limits_{j}a_j\mathbf{I}$, we have that 
	$$	\|\hat{\bm{\beta}}\|^2\leq\frac{1}{\min\limits_{j}a_j^2}(\bm{\epsilon^T\mathbf{X}\mathbf{X}^T\bm{\epsilon}}+{\bm{\beta}^*}^T\mathbf{X}^T\mathbf{X}\mathbf{X}^T\mathbf{X}\bm{\beta}^*+2{\bm{\beta}^*}^T\mathbf{X}^T\mathbf{X}\mathbf{X}^T\bm{\epsilon})$$ and furthermore,
	\begin{equation*}
		\begin{aligned}
			E\|\hat{\bm{\beta}}\|^2\leq & \frac{\sum\limits_{j}a_j}{\min\limits_{j}a_j^2}+\frac{\sum\limits_{j}a_j^2\beta_j^2}{\min\limits_{j}a_j^2}
			\leq \frac{p\kappa}{\min\limits_{j}a_j}+\|\bm\beta^*\|^2\kappa^2.
		\end{aligned}
	\end{equation*}

	So 
$$
			\log(E\|\hat{\bm{\beta}}\|^2)^{1/2}
			\leq\frac{1}{2}\log(\frac{p\kappa^2}{\min\limits_{j}a_j}+\kappa^2\|\bm{\beta}^*\|^2)\leq\log(\kappa)+\frac{1}{2}\log(\frac{p}{\min\limits_{j}a_j}+\|\bm{\beta}^*\|^2).
$$
	By Lemma \ref{En},
	\begin{equation*}
		\begin{aligned} &(E\|\hat{\bm{\beta}}\|^2)^{1/2}(\mathbb{P}(E^c))^{1/2}\\
			\leq &
			\exp(-\frac{1}{4}(\sqrt{\alpha}+\sqrt{\alpha-1})^2+\frac{1}{2}\log(p-q)+\log(\kappa)+\frac{1}{2}\log(\frac{p}{\min\limits_{j}a_j}+\|\bm{\beta}^*\|^2))\\
		\end{aligned}
	\end{equation*}
	and
	$$\|\bm{\beta}^*\|\mathbb{P}(E^c)\leq \exp(-\frac{1}{2}(\sqrt{\alpha}+\sqrt{\alpha-1})^2+\log(p-q)+\log(\|\bm{\beta}^*\|)).$$

	From Lemma \ref{Fevent},
	we also get
	\begin{equation*}
		\begin{aligned} &(E\|\hat{\bm{\beta}}\|^2)^{1/2}(\mathbb{P}(F^c))^{1/2}\\
			\leq &
			\exp(-\frac{1}{4}((1-\eta)\min\limits_{j\in \mathcal{Q}}a_j^{1/2}|\beta_j^*|-(\sqrt{\alpha}+\sqrt{\alpha-1}))^2+\frac{1}{2}\log(q)+\log(\kappa)+\frac{1}{2}\log(\frac{p}{\min\limits_{j}a_j}+\|\bm{\beta}^*\|^2))\\
			&+\exp(-\frac{1}{4}(\min\limits_{j\in \mathcal{Q}}a_j^{1/2}|\beta_j|-(\sqrt{\alpha}+\sqrt{\alpha-1}))^2+\frac{1}{2}\log(q)+\log(\kappa)+\frac{1}{2}\log(\frac{p}{\min\limits_{j}a_j}+\|\bm{\beta}^*\|^2))
		\end{aligned}
	\end{equation*}
	and
	\begin{equation*}
		\begin{aligned}
			&\|\bm{\beta}^*\|\mathbb{P}(F^c)\\
			\leq&
			\exp(-\frac{1}{2}((1-\eta)\min\limits_{j\in \mathcal{Q}}a_j^{1/2}|\beta_j^*|-(\sqrt{\alpha}+\sqrt{\alpha-1}))^2+\log(q)+\log(\|\bm\beta^*\|))\\
			&+\exp(-\frac{1}{2}(\min\limits_{j\in \mathcal{Q}}a_j^{1/2}|\beta_j|-(\sqrt{\alpha}+\sqrt{\alpha-1}))^2+\log(q)+\log(\|\bm\beta^*\|))
		\end{aligned}
	\end{equation*}
	when $(1-\eta)\min\limits_{j\in \mathcal{Q}}a_j^{1/2}|\beta_j^*|-(\sqrt{\alpha}+\sqrt{\alpha-1})\geq 0$. Thus we get the final conclusion.
\end{proof}